\documentclass[10pt,twocolumn,letterpaper]{article}

\usepackage{cvpr}
\usepackage{times}
\usepackage{epsfig}
\usepackage{graphicx}
\usepackage{amsmath}
\usepackage{amssymb}
\usepackage{color}
\usepackage{subfigure}
\usepackage{tabularx}
\usepackage{multirow}
\usepackage{wasysym}
\newcolumntype{C}[1]{>{\centering\let\newline\\\arraybackslash\hspace{0pt}}p{#1}}

\usepackage[pagebackref=true,breaklinks=true,letterpaper=true,colorlinks,bookmarks=false]{hyperref}

\cvprfinalcopy % *** Uncomment this line for the final submission

\def\cvprPaperID{1357} % *** Enter the CVPR Paper ID here

\ifcvprfinal\fi
\begin{document}

%%%%%%%%% TITLE
\title{Learning Transferrable Knowledge for Semantic Segmentation \\ with Deep Convolutional Neural Network}

\author{
Seunghoon Hong$^{\dagger,\ddagger}$\hspace{1.25cm} Junhyuk Oh$^\ddagger$\hspace{1.25cm} Bohyung Han$^\dagger$\hspace{1.25cm} Honglak Lee$^\ddagger$\\
%Seunghoon Hong$^\dagger$, Junhyuk Oh$^\ddagger$, Bohyung Han$^\dagger$, Honglak Lee$^\ddagger$\\
%\normalsize{
\hspace{-0.7cm}
\begin{tabular}{c c}
$^\dagger$Dept. of Computer Science and Engineering & $^\ddagger$Dept. of Electrical Engineering and Computer Science \\
POSTECH, Pohang, Korea & University of Michigan, Ann Arbor, MI, USA \\
{\tt\small \{maga33,bhhan\}@postech.ac.kr} & {\tt\small \{junhyuk,honglak\}@umich.edu}
\end{tabular}
%}
}

\maketitle
%\thispagestyle{empty}

% ====================================================================
% Abstract
% !TEX root = cvpr2016_transfer.tex
\begin{abstract}
We propose a novel weakly-supervised semantic segmentation algorithm based on Deep Convolutional Neural Network (DCNN). 
%Contrary to existing weakly-supervised semantic segmentation algorithms, our model exploits segmentation annotations from non-relevant categories to guide the segmentation on images with only image-level class labels. 
Contrary to existing weakly-supervised approaches, our algorithm exploits auxiliary segmentation annotations available for different categories to guide segmentations on images with only image-level class labels. 
To make the segmentation knowledge transferrable across categories, we design a decoupled encoder-decoder architecture with attention model. 
In this architecture, the model generates spatial highlights of each category presented in an image using an attention model, and subsequently generates foreground segmentation for each highlighted region using decoder.
%Training is naturally done with images with class-labels for attention model with classification objective function, and images with segmentations for decoder.
%The model is natural to incorporate hybrid annotations from different categories, where the decoder is trained with images with segmentation masks and attention model is trained using all images with classification objective. 
%The model is natural to incorporate different source of data for training, and does not rely on heuristic assumptions for segmentation typically required for other weakly-supervised approaches.
%Training attention model requires only image-level labels, and the decoder network trained with other semantic categories can be transferred to generate segmentations for weakly-annotated images.
Combining attention model, we show that the decoder trained with segmentation annotations in different categories can boost the performance of weakly-supervised semantic segmentation. 
%Training attention model requires only image-level labels, while the decoder network requires pixel-wise annotations.
%
%We show that information for shape generation encoded in the decoder network is transferable between different categories, and can be used to generate segmentations in 
%Combined with attention model, we show that the decoder network captures shape information transferable between different categories, and can be used to generate segmentations for weakly-annotated images.
%By sharing 
The proposed algorithm demonstrates substantially improved performance compared to the state-of-the-art weakly-supervised techniques in challenging PASCAL VOC 2012 dataset when our model is trained with the annotations in 60 exclusive categories in Microsoft COCO dataset. 
\end{abstract}

% ====================================================================
% Introduction
% !TEX root = cvpr2016_transfer.tex
\section{Introduction}
\label{sec:intro}
%
%\begin{figure}
%\centering
%\vspace{3cm}
%FIGURE describing motivation of this paper  \\
%\vspace{2.5cm}
%\caption{
%Motivation of this paper.
%}
%\label{fig:motivation}
%\end{figure}
%
\ifdefined\paratitle {\color{blue} 
[Introduction of semantic segmentation.] \\
} \fi
Semantic segmentation refers to the task assigning dense class labels to each pixel in image.
Although pixel-wise labels provide richer descriptions of images than bounding box labels or image-level tags, inferring such labels is a much more challenging task as it involves a highly complicated structured prediction problem. 
%Despite of such challenges, recent breakthrough in Convolutional Neural Network (CNN) brought significant progress in the field of semantic segmentation~\cite{}.
%it captures high-level contents of an image through multiple feature hierarchies even with large per-pixel variations, and appropriate to produce structured output since the prediction is based on local patches in the image.

\ifdefined\paratitle {\color{blue} 
[Fully-supervised approaches] \\
} \fi
%Despite of such challenges, recent breakthrough in Convolutional Neural Network (CNN) brought significant progress in the field of semantic segmentation~\cite{}.
Recent breakthrough in semantic segmentation has been mainly accelerated by the approaches based on Convolutional Neural Networks (CNNs)~\cite{Fcn,Deeplabcrf,Hypercolumns,Sds,Zoomout}.
Given a classification network pre-trained on a large image collection, they learn a network for segmentation based on strong supervision---pixel-wise class labels.
%It captures high-level contents of an image through multiple feature hierarchies even with large per-pixel variations, and produces structured output based on local patches in the image.
%The standard training pipeline is to convert classification network pre-trained on a large image collection to segmentation network, 
%In these approaches, the network is trained with ground-truth segmentations to produce pixel-wise class label predictions. 
%However, similar to other CNN-based approaches, the prediction quality of these approaches is heavily depended on both quantity and quality of training data--in this case, the pixel-wise annotations.
%However, training CNN requires a large amount of fine-quality segmentation annotations.
%Since collecting such data requires extensive labeling costs, the generalizability of these approaches to large-scale problem is significantly limited in practice.
Although the approaches substantially improve the performance over the prior arts, training CNN requires a large number of fine-quality segmentation annotations, which are difficult to collect due to extensive labeling cost.
%For this reason, the generalizability of these approaches to various categories has been significantly limited in practice.
For this reason, scaling up the semantic segmentation task to a large number of categories  is very challenging in practice.
%training CNN typically requires a large number of pixel-wise annotations, which limits the generalizability of the method to various categories in practice. 
%They are suffer from collecting training data in both quantity and quality. main bottleneck is collecting sufficient training data.

\ifdefined\paratitle {\color{blue} 
[Semi- and Weakly- supervised approaches] \\
} \fi
%Limitations: Very weak assumptions on class label. there is no prior or constraints on object shape, location, etc.
Weakly-supervised semantic segmentation~\cite{Boxsup,Wsl,Wssl,Fcmil} is an alternative approach to alleviate annotation efforts in supervised methods.
%Given an weak annotations such as image-level class labels or bounding box annotations, they directly infer the latent segmentation labels from training images based on Multiple Instance Learning (MIL) framework~\cite{} or iterative refinemet procedures~\cite{}.
They infer latent segmentation labels from training images given weak annotations such as image-level class labels~\cite{Wsl,Wssl,Fcmil} or bounding boxes~\cite{Boxsup}.
Since such annotations are easy to collect and even already available in existing datasets~\cite{Imagenet}, it is straightforward to apply those approaches to large-scale problems with many categories.
However, the segmentation quality by the weakly-supervised techniques is typically much worse than the one by supervised methods since there is no direct supervision for segmentation such as object shapes and locations during training.

\ifdefined\paratitle {\color{blue} 
[Our approach: intuition and problem setting] \\
} \fi
The objective of this paper is to reduce the gap between semantic segmentation algorithms based on strong supervisions (\eg, semi- and fully-supervised approaches) and weak supervisions (\eg, weakly-supervised approaches).
Our key idea is to employ segmentation annotations available for different categories to compensate for missing supervisions in weakly annotated images.
%No additional cost is required to collect such data since there are already several datasets publicly available with pixel-wise annotations, \eg, BSD~\cite{MartinFTM01}, Microsoft COCO~\cite{Mscoco}, and LabelMe~\cite{torralbaRY10}, which have not been actively explored yet for semantic segmentation {\color{red}due to variety of categories they are composed of.}
No additional cost is required to collect such data since there are already several datasets publicly available with pixel-wise annotations, \eg, BSD~\cite{MartinFTM01}, Microsoft COCO~\cite{Mscoco}, and LabelMe~\cite{torralbaRY10}.
These datasets have not been actively explored yet for semantic segmentation due to the mismatches in semantic categories with the popular benchmark datasets, \eg PASCAL VOC~\cite{Pascalvoc}.
The critical challenge in this problem is to learn common prior knowledge for segmentation transferrable across categories. 
It is not a trivial task with existing architectures, since they simply pose the semantic segmentation as pixel-wise classification and it is is difficult to exploit examples from the unseen classes.
 %Our key idea is to exploit the segmentation annotations generally available for other categories.
%Our key idea is to exploit the segmentation annotations available for general categories to learn prior for segmentation, and transfer this knowledge to weakly-annotated images to improve the weakly-supervised semantic segmentation.
%Our key idea is to fill the missing information for segmentation in weakly-supervised approaches using segmentation annotations from other categories.
%Exploiting segmentation annotations from other categories is practically more useful, since they are already available in many benchmark datasets (e.g. VOC, MSCOCO).
%This is practically more useful since there are many segmentation annotations available in general. 
%It is not a trivial task with existing architectures since they simply pose the semantic segmentation as region-based classification problem. 
%{\color{red}[Justification is weak.]}

\ifdefined\paratitle {\color{blue} 
[Our approach: details] \\
} \fi
%We propose a novel encoder-decoder architecture based on attention model, which is conceptually appropriate to capture category-independent segmentation knowledge transferable between different categories.
We propose a novel encoder-decoder architecture with attention model, which is conceptually appropriate to transfer segmentation knowledge from one category to another.
In this architecture, the attention model generates \textit{category-specific} saliency on each location of an image, 
while the decoder performs foreground segmentation using the saliency map based on \textit{category-independent} segmentation knowledge.
Our model trained on one dataset is transferable to another by adapting the attention model to focus on  unseen categories.
Since the attention model is trainable with only image-level class labels, our algorithm is applicable to semantic segmentation on weakly-annotated images through transfer learning.
The contributions of this paper are summarized below.
\begin{itemize}
\item 
%We propose a novel approach for weakly-supervised semantic segmentation, which exploits  segmentation annotations from different categories to guide segmentations with weak annotations. 
We propose a new paradigm for weakly-supervised semantic segmentation, which exploits  segmentation annotations from different categories to guide segmentations with weak annotations. 
To our knowledge, this is the first attempt to tackle the weakly-supervised semantic segmentation problem by transfer learning.
\item We propose a novel encoder-decoder architecture with attention model, which is  appropriate to transfer the segmentation knowledge across categories.
%category-independent segmentation knowledge.
\item The proposed algorithm achieved substantial performance improvement over existing weakly-supervised approaches with segmentation annotations in exclusive categories.
\end{itemize}

The rest of the paper is organized as follows.
We briefly review related work and introduce our algorithm in Section~\ref{sec:relatedwork} and \ref{sec:overview}, respectively. 
The detailed configuration of the proposed network is described in Section~\ref{sec:architecture}.
Training and inference procedures are presented in Section~\ref{sec:trainandinf}. 
Section~\ref{sec:experiments} illustrates experimental results on a benchmark dataset.

% ====================================================================
% Related Work
% !TEX root = cvpr2016_transfer.tex
\section{Related Work}
\label{sec:relatedwork}

\ifdefined\paratitle{\color{blue}
[Supervised DNNs for semantic segmenation]
\\}\fi
%Recent progress in semantic segmentation are mainly driven by supervised approaches based on Convolutional Neural Network (CNN)~\cite{Fcn,Deeplabcrf,Hypercolumns,Sds,Zoomout}.
Recent success in CNN has brought significant progress on semantic segmentation in the past few years~\cite{Fcn,Deeplabcrf,Hypercolumns,Sds,Zoomout}.
By posing the semantic segmentation as region-based classification problem, they train the network to produce pixel-wise class labels using segmentation annotations as training data~\cite{Fcn,Hypercolumns,Sds,Zoomout}.
Based on this framework, some approaches improve segmentation performance by learning deconvolution network to capture accurate object boundaries~\cite{deconvnet} or adopting fully connected CRF as post-processing~\cite{Crfrnn,Deeplabcrf}.
However, the performance of the supervised approaches depends heavily on the size and quality of training data, which limits the scalability of the algorithms.

\ifdefined\paratitle{\color{blue}
[Weakly- and Semi-supervised DNNs to mitigate limitations of fully-supervised approach]
\\}\fi
To reduce the efforts for annotations, weakly-supervised approaches attempt to learn the model for semantic segmentation only with weak annotations~\cite{Wsl,Wssl,Fcmil,Boxsup}.
To infer latent segmentation labels, they often rely on the techniques such as Multiple Instance Learning (MIL)~\cite{Wsl, Fcmil} or Expectation-Maximization (EM)~\cite{Wssl}.
Unfortunately, they are not sufficient to make up missing supervision and lead to significant performance degradation compared to fully-supervised approaches.
In the middle, semi-supervised approaches~\cite{Wssl, decouplednet} exploit a limited number of strong annotations to reduce performance gap between fully- and weakly-supervised approaches.
%Alternatively, semi-supervised approaches~\cite{Wssl, decouplednet} try to reduce the gap between fully- and weakly-supervised approaches using limited number of strong annotations.
Notably, \cite{decouplednet} proposed a decoupled encoder-decoder architecture for segmentation, where it divides semantic segmentation into two separate problems---classification and segmentation---and learns a decoder to perform binary segmentation for each class identified in the encoder.
Although this semi-supervised approach improves performance by sharing the decoder for all classes, it still needs strong annotations in the corresponding classes for segmentation.
We avoid this problem by using segmentation annotations available for other categories.

\ifdefined\paratitle{\color{blue}
[Transfer learning and domain adaptation in other vision problems]
\\}\fi
In the field of computer vision, the idea of employing external data to improve performance of target task has been explored in context of domain adaptation~\cite{Saenko10,Khosla12,Gopalan11,Ganin11} or transfer learning~\cite{Lim11,Tommasi14}. 
However, the approaches in domain adaptation often assume that there are shared categories across domains, and the techniques with transfer learning are often limited to simple tasks such as classification.
We refer \cite{Patricia14} for comprehensive surveys on domain adaptation and transfer learning.
Hoffman~\etal~\cite{Lsda} proposed a large-scale detection system by transferring knowledge for object detection between categories.
Our work shares the motivations with this work, but aims to solve a highly complicated structured prediction problem, semantic segmentation.
%{\color{red} +Visual attention [Want to add discussion about visual attention?]}

\ifdefined\paratitle{\color{blue}
[Attention model in other computer vision problem]
\\}\fi
There has been a long line of research on learning visual attention~\cite{Larochelle10, Bazzani11, Alexe12, Minh15, Ba15, showatttell, Tang14}. 
%They aim to identify salient part of the image or video based on low-level visual characteristics.
Their objective is to learn the attention mechanism that can adaptively focus on salient part of an image or video for various computer vision tasks, such as object recognition~\cite{Larochelle10, Alexe12, Ba15}, object tracking~\cite{Bazzani11}, caption generation~\cite{showatttell}, image generation~\cite{Tang14}, etc.
Our work is an extension of this idea to semantic segmentation by transfer learning.
\section{Algorithm Overview}
\label{sec:overview}

\iffalse
\begin{table} \footnotesize
\centering
\caption{Comparison between different problem settings.} \vspace{0.2cm}
\begin{tabular}
{
@{}C{2.3cm}@{}|@{}C{1.5cm}@{}|@{}C{1.8cm}@{}|@{}C{2.5cm}@{}
}
\hline
Method & Class label & Segmentation & Extra annotation \\
\hline
Transfer (Ours) & $\ocircle$ & $\times$ & segmentation from other classes \\
\hline
Weakly-supervised & $\ocircle$ & $\times$ & prior knowledge \\
% MIL & $\ocircle$ &  & \\
% CCNN & $\ocircle$ & & \\
% EM-Adapt & $\ocircle$ & & \\
\hline
Semi-supervised & $\ocircle$ & $\triangle$ &  \\
\hline
Fully-supervised & $\ocircle$ & $\ocircle$ &  \\
\hline
\end{tabular}
\label{tab:problem_setting}
\end{table}
\fi

\begin{figure*}[!t]
\centering
% \vspace{2cm}
\includegraphics[width=0.95\linewidth] {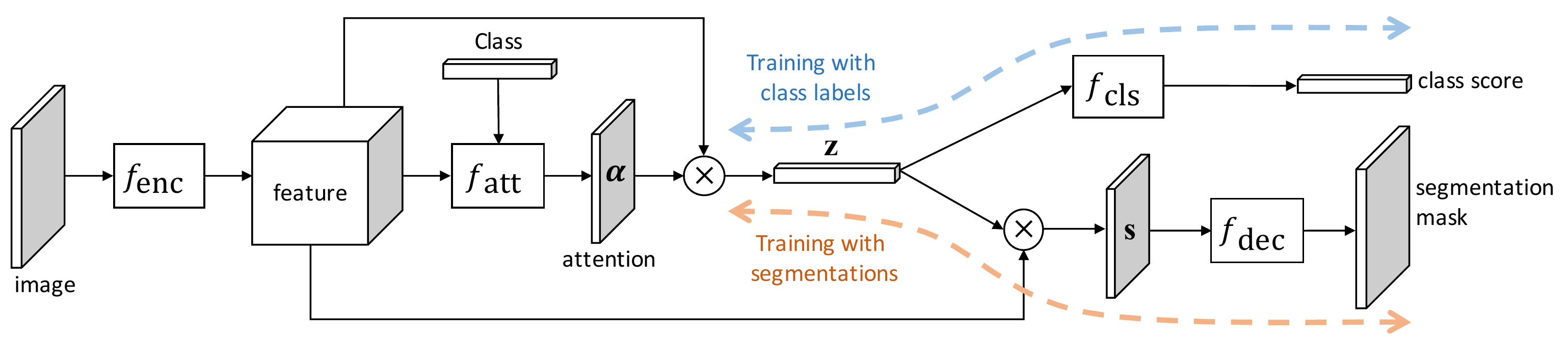}
% Figure describing overall architecture of the proposed algorithm
% \vspace{2cm}
\caption{
Overall architecture of the proposed algorithm. 
%Given a feature extracted from encoder, the attention model predicts salient regions of individual classes presented in the image. 
Given a feature extracted from the encoder, the attention model estimates adaptive spatial saliency of each category associated with input image (Section~\ref{sec:attention}).
The outputs of attention model are subsequently fed into the decoder, which generates foreground segmentation mask of each focused region (Section~\ref{sec:segmentation}).
% During training, we fix the encoder with pre-trained weights, and jointly optimize both the decoder and the attention model using the images with pixel-wise annotations (in source domain) and {\color{red}class labels} (in both source and target domains), respectively.
During training, we fix the encoder by pre-trained weights, and leverage the segmentation annotations from source domain to train both the decoder and the attention model, and image-level class labels in both domains to train the attention model. 
After training, semantic segmentation on the target domain is performed naturally by exploiting the decoder trained with source images and the attention model adapted to target domain (Section~\ref{sec:trainandinf}). 
%{\color{red}[comment about image: why `Class' and `'class score' are different?]}
%{\color{red} [How to train encoder? Just transfer without fine-tuning?]}
%which is the natural process of transferring the segmentation knowledge encoded in the decoder to the target domain.
%After training, semantic segmentation on the target domain can be naturally performed by feed-forwarding images until decoder, which is the natural process of transferring the segmentation knowledge encoded in the decoder to the target domain.
% learn the attention model using images with class labels to maximize classification performance, and the decoder using images with segmentation annotations to maximize segmentation performance.
%By training the attention model with images in both target and source domains using weak annotations to maximize classification performance, we 
%With this architecture, we train the attention model with images in both target and source domain using weak annotations to maximize classification performance, and train the decoder with images in source domain using pixel-wise annotations. 
%The output of attention model is used to subsequently generate features for classification and segmentation, which are optimized with images in target and source domains, respectively.
}
\label{fig:overview}
\end{figure*}

\ifdefined\paratitle {\color{blue} 
[Descriptions of the problem setting] \\
} \fi
%\B{$\mathcal{T}=\left\{\left(\textbf{x}_{i}, \left\{\textbf{y}^l_i\right\}^{\matchal{L}^*_{i}}_{l}\right)\right\}^{N_t}_{i}$}
%and
%\B{$\mathcal{S}=\left\{\left(\textbf{x}_{i}, \left\{\textbf{y}^l_i,\textbf{d}^l_i\right\}^{\matchal{L}^*_{i}}_{l}\right)\right\}^{N_s}_{i}$}

This paper tackles the weakly-supervised semantic segmentation problem in transfer learning perspective. 
Suppose that we have two sources of data, $\mathcal{T}=\{1,...,N_t\}$ and $\mathcal{S}=\{1,...,N_s\}$, which are composed of $N_t$ and $N_s$ images, respectively.
Note that a set of images in \textit{target} domain, denoted by $\mathcal{T}$, only have image-level class labels while the other set of data in \textit{source} domain, referred to as $\mathcal{S}$, have pixel-wise segmentation annotations.
%We will refer to the data with image-level class labels as \textit{target} domain, and the data with pixel-wise annotations as \textit{source} domain. 
Our objective is to improve the weakly-supervised semantic segmentation on the target domain using the segmentation annotations available in the source domain. 
%Note that both domains may be associated with different set of categories. 
%In this paper, we assume the extreme case where two domains have exclusive set of categories.
We assume that both target and source domains are composed of exclusive sets of categories.
%\footnote{Although it may be more reasonable to assume some shared categories in practice, we assume the extreme case to demonstrate capability of the proposed algorithm.}
In this setting, there is no direct supervision (\ie, ground-truth segmentation labels) for the categories in the target domain, which makes our objective similar to a weakly-supervised semantic segmentation setting.
%Since our algorithm captures category-independent segmentation knowledge, any datasets with pixel-wise annotations can be employed as source domain data.

\ifdefined\paratitle {\color{blue} 
[Detailed description of the proposed architecture] \\
} \fi
%To capture the category-independent segmentation knowledge from source domain, and transfer the knowledge to the target domain, we design a decoupled encoder-decoder network based on attention model.
To transfer segmentation knowledge from source to target domain, we propose a novel encoder-decoder architecture with attention model.
Figure \ref{fig:overview} illustrates the overall architecture of the proposed algorithm. 
The network is composed of three parts: encoder, decoder and attention model between the encoder and the decoder.
%that convert outputs from the encoder to more general and meaningful representation to decoder.
In this architecture, the input image is first transformed to a multi-dimensional feature vector by the encoder, and the attention model identifies salient region for each category associated with the image.
The output of the attention model reveals location information of each category in a coarse feature map, where the dense and detailed foreground segmentation mask for each category is subsequently obtained by the decoder.

\ifdefined\paratitle {\color{blue} 
[Descriptions about training procedure] \\
} \fi
Training our network involves different mechanisms for source and target domain examples, since they are associated with heterogeneous annotations with different levels of supervision.
We leverage the segmentation annotations from source domain to train both the decoder and the attention model with segmentation objective, while image-level class labels in both target \textit{and} source domains are used to train the attention model under classification objective.
\iffalse
Given examples from source domain, we train both the decoder and the attention model to maximize the segmentation performance.
Then given weak annotations (\ie, image-level class labels) in target $\textit{and}$ source domains, we train the attention model to maximize the classification performance.
\fi
% by putting additional layers for classification on top of attention model.
The training is performed jointly for both objectives using examples from both domains. 

\ifdefined\paratitle {\color{blue} 
[Intuitive descriptions on advantage of our method] \\
} \fi
The proposed architecture exhibits several advantages to capture transferrable segmentation knowledge across domains.
Employing the decoupled encoder-decoder architecture~\cite{decouplednet} makes it possible to share the information for shape generation among different categories. %, since the objective of the decoder is figure-ground segmentation.
The attention model provides not only predictions for localization but also category-specific information that enables us to adapt the decoder trained in source domain to target domain. 
%Attention model provides not only predictions for localization but also category-independent representation that allows the decoder trained with source domain can be used in segmentation of the target domain. %that can be used between different categories.
The combination of two components makes information for segmentation transferable across different categories, and provides useful segmentation prior that is missing in weakly annotated images in target domain.
%The attention model and decoupled decoder architecture play a key role in transferring the segmentation knowledge between different domains.
%Our architecture has two separate branches for classification and segmentation. 
%The attention model builds a representation for both branches given a feature extracted from the encoder. 
%With this architecture, we can train the attention model using either images with class labels for classification, and pixel-wise annotations for segmentation. 
%Given this architecture, the inference for the new image is performed as follows. 

% ====================================================================
% Algorithm Details
% !TEX root = cvpr2016_transfer.tex

\section{Architecture}
\label{sec:architecture}
This section describes the architecture of the proposed algorithm, including the attention 
model and the decoder.

\subsection{Preliminaries}
We first describe notations and general configurations of the proposed model.
Our network is composed of four parts, $f_{\text{enc}}, f_{\text{att}}, f_{\text{cls}}$ and $f_{\text{dec}}$, which are neural networks corresponding to encoder, attention model, classifier and decoder, respectively. 
Our objective is to train all components using the examples from both domains.

Let ${\bf x}$ denotes a training image from either source or target domain.
We assume that the image is associated with a set of class labels $\mathcal{L}^*$, which is given by either ground-truth (in training) or prediction (in testing).
Given an input image ${\bf x}$, the network first extracts a feature descriptor as
\begin{equation}
{\bf A} = f_{\text{enc}}({\bf x}; \theta_e), ~~~ {\bf A} \in \mathrm{R}^{M\times D}
\label{eqn:feature}
\end{equation}
where $\theta_e$ is the model parameter for the encoder, and $M$ and $D$ denote the number of hidden units in each channel and the number of channels, respectively.
We employ VGG-16 layer net~\cite{Vgg16} pre-trained on ImageNet~\cite{Imagenet} as our encoder $f_{\text{enc}}$, and the feature descriptor ${\bf A}$ is obtained from the last convolutional layer to retain spatial information in input image.
%We will omit the subscript in following sections for brevity.
The extracted feature and associated labels are then used to generate attention, which is discussed in the following subsection.

\subsection{Attention model}
\label{sec:attention}
\begin{figure}
\centering
%\vspace{1cm}
%Input image,  attentions given different class labels\\
%\vspace{1cm}
\subfigure[Input image]{
\includegraphics[width=0.24\linewidth] {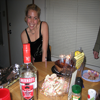} 
}\hspace{-0.21cm}
\subfigure[${\boldsymbol \alpha}^{bottle}$]{
\includegraphics[width=0.24\linewidth] {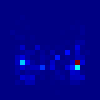} 
}\hspace{-0.21cm}
\subfigure[${\boldsymbol \alpha}^{table}$]{
\includegraphics[width=0.24\linewidth] {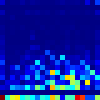} 
}\hspace{-0.21cm}
\subfigure[${\boldsymbol \alpha}^{person}$]{
\includegraphics[width=0.24\linewidth] {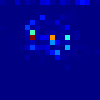}
}
\caption{
Examples of learned attentions.  (a) Input image, (b), (c) and (d) represent attention weights obtained by Eq.~\eqref{eqn:attention}. 
The proposed attention model adaptively focuses on different areas in an image depending on input labels.
}
\label{fig:attention}
\end{figure}
%
%This section describes the attention model, which identifies salient part of an image for segmentation and is the key to adapt the learned model to different domains.
\ifdefined\paratitle {\color{blue} 
[Attention model $f_{\text{att}}$ to generate attention] \\
} \fi
%We follow the similar principles employed in \cite{} to model our attention.
Given a feature descriptor extracted from the encoder ${\bf A}\in\mathrm{R}^{M\times D}$ and its associated class labels $\mathcal{L}^*$, the objective of our attention model is to learn a set of positive weight vectors $\{ {\boldsymbol \alpha}^l \}_{\forall l\in\mathcal{L}^*}$ defined over a 2D space, where each element of ${\boldsymbol \alpha}^l\in\mathrm{R}^M$ represents the relevance of each feature location to $l^{th}$ category.
%Following formulations in \cite{showatttell},
Our attention model can be formally given by
%Following the similar principles employed in \cite{showatttell}, our model generates the attention by
%
\begin{align}
{\bf v}^l &= f_{\text{att}}({\bf A}, {\bf y}^l; \theta_\alpha), ~~~{\bf v}^l\in\mathrm{R}^M\\
\alpha_i^{l} &= \frac{\exp \left( v_i^l \right) }{\sum_i \exp \left( v_i^l \right)}, ~~~~~{\boldsymbol \alpha}^{l}\in\mathrm{R}^M,
\label{eqn:attention}
\end{align}
where ${\bf y}^l$ denotes a one-hot label vector for the $l^{th}$ category, and ${\bf v}^l$ represents unnormalized attention weights. %$\theta_\alpha$ denotes parameters associated with the attention model $f_{\text{att}}$.
To encourage the model to pay attention to only a part of the image, we normalize ${\bf v}^l$ to ${\boldsymbol \alpha}^l$ using a softmax function as suggested in \cite{showatttell}.

\ifdefined\paratitle {\color{blue} 
[Detailed structure of  $f_{\text{att}}$] \\
} \fi
To generate category-specific attention ${\boldsymbol \alpha}^l$ using our attention model $f_{\text{att}}$, we employ multiplicative interactions~\cite{Multiplicative} between feature and label vector.
It learns a set of gating parameters represented by a 3-way tensor to model correlation between feature and label vectors. %${\boldsymbol \alpha}^l$ and ${\bf y}^l$.
For scalability issue, we reduce the number of parameters by the factorization technique proposed in \cite{Multiplicative}, and our model can be written as
%We observe that using multiplicative interaction generally gives better results than additive interactions (e.g. concatenation), probably because it is more appropriate to capture correlation between feature and label.
%Due to huge dimensions of feature ${\bf a}$, we employed factorized version proposed in \cite{Multiplicative}, which is 
%
\begin{equation}
{\bf v}^l = \mathbf{W}^{{att}}\left( \mathbf{W}^{{feat}} {\bf A} \odot \mathbf{W}^{{label}} {\bf y}^l  \right) + {\bf b},
\label{eqn:multiplicative}
\end{equation}
where $\odot$ denotes element-wise multiplication and ${\bf b}\in\mathrm{R}^M$ is bias.
Note that the weights are given by ${\bf W}^{att}\in\mathrm{R}^{d\times MD}, {\bf W}^{label}\in\mathrm{R}^{d\times L}$ and ${\bf W}^{att}\in\mathrm{R}^{M\times d}$, where $L$ and $d$ denote the size of label vector and the number of factors, respectively.
We observe that using multiplicative interaction generally gives better results than additive ones (\eg, concatenation), because it is capable of capturing high-order dependency between the feature and the label.
%probably because it is more appropriate to capture correlation between feature and label.

\ifdefined\paratitle {\color{blue} 
[Training attention with weakly-annotated images] \\
} \fi
To apply the attention to our transfer-learning scenario, the model $f_{\text{att}}$ should be trainable in both target and source domains.
Since examples from the target domain are associated with only image-level class labels, 
we put additional layers $f_{\text{cls}}$ on top of the attention model, and optimize both $f_{\text{att}}$ and $f_{\text{cls}}$ under classification objective.
To this end, we extract features based on the category-specific attention by aggregating features over the spatial region as follows: % we generate another representation of the input image by combining the feature and category-specific attention by 
%To this end, we generate another representation of the input image by encoding class-specific spatial information in ${\bf \alpha}^l$ to original feature map ${\bf a}$ by
%The attention model in Eq.~\eqref{eqn:attention} is optimized under classification objective.
%To this end, we first encode spatial information obtained by attention to original feature map by
%
\begin{equation}
{\bf z}^l =   {\bf A}^\mathrm{T}{\boldsymbol \alpha}^l , ~~~~{\bf z}^l \in \mathrm{R}^D.
\label{eqn:context_vector}
\end{equation}
Intuitively, ${\bf z}^l$ represents a category-specific feature defined over all the channels in the feature map.
%It can also be interpreted as a bag-of-words representation where the vocabulary consists of convolution filters. 
%Each vector ${\bf z}^l$ corresponds to a dynamic representation of an image with respect to the $l$th category.

Using the weak annotations associated with both target and source domain images, our attention model is trained to minimize the classification loss as follows:
%Then using the weak annotations associated with both target and source domain images, we learn our attention model to maximize the classification performance by
%
\begin{equation}
\min_{\theta_\alpha, \theta_c} \sum_{i\in \mathcal{T}\cup\mathcal{S}} \sum_{l\in\mathcal{L}^*_i} e_c\left({\bf y}_i^l, f_{\text{cls}} ({\bf z}_i^l;\theta_c) \right),
\label{eqn:obj_cls}
\end{equation}
where $e_c$ denote the loss between ground-truth ${\bf y}_i^l$ and predicted label vector $f_{\text{cls}} ({\bf z}_i^l;\theta_c)$, respectively, which are both defined for single label $l$ and aggregated over $\forall l \in \mathcal{L}^*_l$ for each image.
We employed softmax loss function to measure classification loss $e_c$.

\iffalse
where $\mathcal{L}^*_i$ denote one-hot label vector associated with $i$th image, and $f_{\text{cls}}$ is multi-layer perceptron for classification that taking encoded feature ${\bf z}$ as an input, and generate label predictions, and $\theta_c$ denote associated parameters with $f_{\text{cls}}$.
Since the objective function in Eq.~\eqref{eqn:obj_cls} can be optimized with weak-annotation, we can train the attention model for categories in the target domain.
\fi

\ifdefined\paratitle {\color{blue} 
[Discussions for label to label prediction] \\
} \fi
The optimization of Eq.~\eqref{eqn:obj_cls} may involve potential overfitting issue, 
%since it uses the same label vector ${\bf y}^l$ in both attention generation and classification.
since the ground-truth class label vector ${\bf y}^l$ is given as an input to attention model as well.
In practice, we observe that our model can avoid this issue by effectively eliminating the direct link from attention to label prediction and constructing intermediate representation ${\bf z}$ based on the original feature ${\bf A}$. 

\iffalse
The attention model described above may has potential over-fitting issue, since it basically takes label as an input, and aims to estimate the same label as output of classification network.
We observe that our model can avoid this issue effectively, probably because our model constructs intermediate representation ${\bf z}$ as input to decoder, not directly using the attention itself.
Since there is no direct link from attention to decoder, it prevents the model not overfit to trivial label-to-label link ignoring information from the feature.
\fi

%Figure~\ref{fig:attention} describes examples of learned attention weights for each class.
Figure~\ref{fig:attention} illustrates the learned attention weights for individual classes. 
We observe that the attention model adapts spatial saliency effectively depending on its input labels.

%%%%%%%%%%%%%%%%%%%%%%%%%%%%%%%%%%%%%%%%%%%
%%%%%%%%%%%%%%%%%%%%%%%%%%%%%%%%%%%%%%%%%%%
\subsection{Decoder}
\label{sec:segmentation}
\ifdefined\paratitle {\color{blue} 
[Transition: issues on directly using attention for segmentation] \\
} \fi
The attention model described in the previous section generates a set of adaptive saliency maps for each category $\{ {\boldsymbol \alpha}^l \}_{\forall l \in \mathcal{L}^*}$, which provides useful information for localization.  %reveals useful information for localization.
%\B{However, the attention weights may not be detailed enough to be directly used as input for the decoder, because they tend to be sparse and capture only a key part of an object (as shown in Figure~\ref{fig:attention}) due to the softmax operation in Eq.~\ref{eqn:attention}, which encourages the model to focus on only a small part of the image that is useful for classification.} 
Given these attentions, the next step of our algorithm is to reconstruct dense foreground segmentation mask for each attended category by the decoder.
However, the direct application of attentions weights to segmentation may be problematic, since the activations tend to be sparse due to the softmax operation in Eq.~\eqref{eqn:attention} and  may lose information encoded in the feature map useful for shape generation.

\begin{figure}
\centering
%\vspace{1cm}
%FIGURE describing densified attention \\
%\vspace{1cm}
\includegraphics[width=0.32\linewidth] {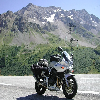} \hspace{-0.12cm}
\includegraphics[width=0.32\linewidth] {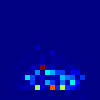} \hspace{-0.12cm}
\includegraphics[width=0.32\linewidth] {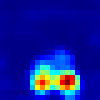} \\
\vspace{-0.15cm}
\subfigure[Input image]{
\includegraphics[width=0.32\linewidth] {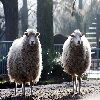}
}\hspace{-0.2cm}
\subfigure[Attention]{
\includegraphics[width=0.32\linewidth] {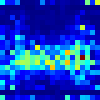}
}\hspace{-0.2cm}
\subfigure[Densified attention]{
\includegraphics[width=0.32\linewidth] {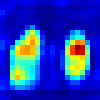}
}
\caption{
Examples of attention (${\boldsymbol \alpha}^l$) and densified attention (${\bf s}^l$).
}
\label{fig:decoder_input}
\end{figure}

\iffalse
Given outputs of the attention model described in Section~\ref{sec:attention}, our next task is to reconstruct dense segmentation mask in original image coordinate by decoder.
The attention outputs in Eq.~\eqref{eqn:attention} reveals coarse location of each category in a feature map, and provide strong prior for segmentation. 
However, directly employing the attention for segmentation may be problematic, since the attention weights tends to be sparse as the attention for different class should be discriminative, and loose useful information in feature map.
\fi
\ifdefined\paratitle {\color{blue} 
[Constructing inputs to decoder (densified attention)] \\
} \fi
%\B{To construct more fine-grained attentions for the decoder, we propose to exploit ${\bf z}^l$ in Eq.~\eqref{eqn:context_vector} by interpreting it as \textit{a channel filter} of the feature map (${\bf A}$). The intuition is that each element (or channel) of ${\bf z}^l$ has a large value when the object in the attention region strongly activates the corresponding channel of the feature map, because ${\bf z}^l$ is simply a summation of activation values over the attention region. 
%Thus, the channels with large values in ${\bf z}^l$ are highly likely to be related to the object that the model is paying attention to.
%Based on this assumption, we compute \textit{densified} attention (${\bf s}^l$) by using ${\bf z}^l$ as a channel filter of the feature map as follows:
%}
To resolve this issue and reconstruct useful information for segmentation, we feed the additional inputs to the decoder using attention ${\boldsymbol \alpha}^l$ and the original feature ${\bf A}$.
Rather then directly using the attention, we exploit the intermediate representation ${\bf z}^l$ obtained from Eq.~\eqref{eqn:context_vector}.
It represents relevance of each \textit{channel} out of the feature maps with respect to $l^\text{th}$ category.
Then we aggregate spatial activations in each channel of the feature using ${\bf z}^l$ as coefficients, which is given by
% Then we generate the input to the decoder by blending spatial activations of each feature channel direction using ${\bf z}^l$ as weight by
%Therefore, we generate the inputs to the decoder given ${\bf a}$ and ${\bf z}^l$ by
%We exploits the context vector ${\bf z}$ obtained from Eq.~\eqref{}, since it reflects the importance of each channel in the feature ${\bf a}$ based on obtained attention weight by
%
\begin{equation}
{\bf s}^l =   {\bf A} {\bf z}^l, ~~{\bf s}\in\mathrm{R}^M
\label{eqn:decoder_input}
\end{equation}
%
%where ${\bf s}^l$ represents reconstructed inputs to the decoder which has same size to attention ${\bf \alpha}^l$.
where ${\bf s}^l$ represents \textit{densified attention} in the same size with ${\boldsymbol \alpha}^l$ and serves as inputs to the decoder.
As shown in Figure~\ref{fig:decoder_input}, densified attention maps preserve more details of the object shape compared to the original attention (${\boldsymbol \alpha}^l$). 
See Appendix~\ref{apdx_sec:densified_att} for more comprehensive analysis of Eq.~\eqref{eqn:decoder_input}.
%More comprehensive analysis of the densified attention is described in Appendix~\ref{apdx_sec:densified_att}.
% Figure~\ref{fig:decoder_input} describes examples of outputs from Eq.~\eqref{eqn:decoder_input}.
%The examples of densified attentions ${\bf s}$ are described in Figure~\ref{fig:decoder_input}.
% We observe that obtained map reveals more dense and detailed configuration of the object, which is more useful to generate object shapes by the decoder.

\ifdefined\paratitle {\color{blue} 
[Decoder objective function] \\
} \fi
Given densified attention ${\bf s}^l$ as input, the decoder is trained to minimize the segmentation loss by the following objective function
% Then given densified attention ${\bf s}^l$ as inputs, we train the decoder to maximize the segmentation performance by
\begin{equation}
\min_{\theta_\alpha, \theta_s} \sum_{i\in \mathcal{S}}  \sum_{l\in\mathcal{L}^*_i} e_s \left( {\bf d}^l_i, f_{\text{dec}}( {\bf s}^l_i;\theta_s)\right),
\label{eqn:obj_seg}
\end{equation}
where ${\bf d}^l_i$ denotes a \textit{binary} segmentation mask of $i^\text{th}$ image for $l^\text{th}$ category, and $e_s$ denotes pixel-wise loss function between ground-truth and predicted segmentation mask. %$\theta_s$ denotes associated parameters to $f_{\text{dec}}$,% 
Similar to classification, we employ a softmax loss function for $e_s$.
Since training requires ground-truth segmentation annotations, the objective function is only optimized with images in source domain.

\ifdefined\paratitle {\color{blue} 
[Decoder architecture] \\
} \fi
We employ recently proposed deconvolution network \cite{deconvnet} for our decoder architecture $f_{\text{dec}}$.
Given an input to the decoder ${\bf s}^l$, it generates a segmentation mask in the same size with input image by multiple successive operations of unpooling, deconvolution and rectification. 
Pooling switches are shared between pooling and unpooling layers, which is appropriate to recover accurate object boundary.
We refer to \cite{deconvnet} for more details.
%Unpooling is implemented by importing the switch variable from every pooling layer in the classification network, and the number of deconvolutional and unpooling layers are identical to the number of convolutional and pooling layers in the classification network.
%We employ the softmax loss function to measure per-pixel loss in Eq.~\eqref{eqn:obj_seg}.

\ifdefined\paratitle {\color{blue} 
[Discussion: shape generation] \\
} \fi
Note that we train our decoder to generate foreground segmentation of each attention ${\boldsymbol \alpha}^l$.
By decoupling classification, which is a domain specific task, from the decoding~\cite{decouplednet}, we can capture category-independent information for shape generation and apply the architecture to any unseen categories.
Since all weights in the decoder are shared between different categories, it potentially encourages the decoder to capture common shape information that can be generally applicable to multiple categories.

%Since the learned deconvolution filters capture global to local object shapes, this sharing such weights may encourage the model to learn common shapes across different categories, which is useful to generate segmentations across domains.

%
%

%%%%%%%%%%%%%%%%%%%%%%%%%%%%%%%%%%%%%%%%%%%
%%%%%%%%%%%%%%%%%%%%%%%%%%%%%%%%%%%%%%%%%%%
\section{Training and Inference}
\label{sec:trainandinf}
\ifdefined\paratitle {\color{blue} 
[Training procedure: using images from both domains jointly.] \\
} \fi
%This section describes the training and inference procedure of the proposed algorithm.
%Our network is trainable with examples from both domains;
%we train the attention model using weak-annotations from both source and target domains (Section~\ref{sec:attention}) and the decoder using pixel-wise annotations from source domain (Section~\ref{sec:segmentation}).
%Given the objective function for attention model and decoder, we train both branch jointly using images from target and source domain.
Combining Eq.~\eqref{eqn:obj_cls} and \eqref{eqn:obj_seg}, the overall objective function is given by
\begin{align}
\min_{\theta_\alpha, \theta_c, \theta_s} &\sum_{i\in \mathcal{T}\cup\mathcal{S}} \sum_{l\in\mathcal{L}^*_i} e_c\left({\bf y}_i^l, f_{\text{cls}} ({\bf z}_i^l;\theta_c) \right) \label{eqn:obj_all}\\\nonumber &+ \lambda \sum_{j\in \mathcal{S}}  \sum_{l\in\mathcal{L}^*_j} e_s \left( {\bf d}^l_j, f_{\text{dec}}( {\bf s}^l_j;\theta_s)\right),
\end{align}
where $\lambda$ controls balance between classification and segmentation.
%During training, we use examples from both domains jointly to optimize attention model and the decoder.
During training, we optimize Eq.~\eqref{eqn:obj_all} using examples from both domains. 
Note that it allows joint optimization of attention model for both classification and segmentation. 
Although our attention model is generally good even trained with only class labels (see Figure~\ref{fig:decoder_input}), training attention based only on classification objective sometimes lead to noisy predictions due to missing supervision for localization. 
By jointly training with segmentation objective, we can regularize it to avoid finding noisy solution for target domain categories.
\iffalse
Training attention solely for classification is similar to solving weakly-supervised learning problem, and may lead to noisy model with bad localization performance.
By learning the attention for segmentation jointly, it may increase the chance to avoid finding noisy attention model for the target domain categories.
\fi
%By learning the attention for segmentation jointly, and it is possible to constrain the model not overfit to bad solution.
%Note that it allows joint optimization of attention model for both classification and segmentation, which is potentially helpful to correct noisy models learned with classification objective by segmentation objective.
%The joint optimization potentially helpful to correct noises in learning attentions for categories in the target domain, since learning attention for classification may not directly enforce to find attentions good for localization, whereas training attention with segmentation directly enforce it.
%
After training, we remove the classification layers $f_{\text{cls}}$ since it is required only in training to learn attentions for the data from target domain categories.

\ifdefined\paratitle {\color{blue} 
[Inference procedure: ]\\
%Remove $f_{\text{cls}}$ after training, and obtain class label vector for separate classification network, iteratively forward propagate decoder and obtain seperate foregorund segmentation mask. how to combine the separate segmentation maps] \\
} \fi
For inference of target domain images with the trained model, we first apply a separate classifier to identify a set of labels $\mathcal{L}^*$ associated with the image.
Then for each identified label $l\in\mathcal{L}^*$, we iteratively construct attention weights ${\boldsymbol \alpha}_i^l$ and obtain foreground segmentation mask $f_{\text{dec}}( {\bf s}^l_i )$ from the decoder output. 
%Then for each identified label $l\in\mathcal{L}^*$, we iteratively construct attention weights ${\bf \alpha}_i^l$ and subsequently obtain foreground segmentation mask from the output of the decoder $f_{\text{dec}}( {\bf s}^l_i )$. 
Given foreground probability maps from all labels $\{f_{\text{dec}}( {\bf s}^l_i )\}_{\forall l\in\mathcal{L}^*}$, the final segmentation label is obtained by taking the maximum probability in channel direction.

%% ====================================================================
% Experiments
% !TEX root = cvpr2016_transfer.tex
\newcommand{\ROT}[1]{{\rotatebox[origin=c]{90}{#1}}}

\section{Experiments}
\label{sec:experiments}
% This section describes detailed information in implementation and experiment, and provides results in a challenging benchmark dataset.
This section describes detailed information about implementation and experiment, and provides results in a challenging benchmark dataset.
%%%%%%%%%%%%%%%%%%%%%%%% INSERT TABLE HERE!
\begin{table*}[!t] \footnotesize
\centering
\caption{Evaluation results on PASCAL VOC 2012 \textit{validation} set.} \vspace{0.1cm}
\begin{tabular}
{
@{}C{2.6cm}@{}|@{}C{0.68cm}@{}C{0.66cm}@{}C{0.66cm}@{}C{0.66cm}@{}C{0.66cm}@{}C{0.66cm}@{}C{0.66cm}@{}C{0.66cm}@{}C{0.66cm}@{}C{0.66cm}@{}C{0.66cm}@{}C{0.66cm}@{}C{0.66cm}@{}C{0.66cm}@{}C{0.75cm}@{}C{0.75cm}@{}C{0.66cm}@{}C{0.66cm}@{}C{0.66cm}@{}C{0.66cm}@{}C{0.66cm}@{}|@{}C{0.75cm}@{}
}
\hline
Method&bkg&aero&bike&bird&boat&bottle&bus&car&cat&chair&cow&table&dog&horse&mbk&person&plant&sheep&sofa&train&tv&mean\\
\hline
\textbf{Weakly-supervised:} \raggedright & & & & & & & & & & & & & & & & & & & & & & \\
% MIL-FCN \raggedright & - & - & - & - & - & - & - & - & - & - & - & - & - & - & - & - & - & - & - & - & - & 24.9 \\ 
% EM-Adapt w/o CRF \raggedright & 65.3 & 28.2 & 16.9 & 27.4 & 21.1 & 28.1 & 45.4 & 40.5 & 42.3 & 13.2 & 32.1 & 23.3 & 38.7 & 32.0 & 39.9 & 31.3 & 22.7 & 34.2 & 22.8 & 37.0 & 30.0 & 32.0 \\ 
EM-Adapt~\cite{Wssl} \raggedright & 67.2 & 29.2 & 17.6 & 28.6 & 22.2 & 29.6 & 47.0 & 44.0 & 44.2 & 14.6 & 35.1 & 24.9 & 41.0 & 34.8 & 41.6 & 32.1 & 24.8 & 37.4 & 24.0 & 38.1 & 31.6 & 33.8 \\ 
% CCNN w/o CRF \raggedright & 66.3 & 24.6 & 17.2 & 24.3 & 19.5 & 34.4 & 45.6 & 44.3 & 44.7 & 14.4 & 33.8 & 21.4 & 40.8 & 31.6 & 42.8 & 39.1 & 28.8 & 33.2 & 21.5 & 37.4 & 34.4 & 33.3 \\ 
CCNN~\cite{Ccnn} \raggedright & 68.5 & 25.5 & 18.0 & 25.4 & 20.2 & 36.3 & 46.8 & 47.1 & 48.0 & 15.8 & 37.9 & 21.0 & 44.5 & 34.5 & 46.2 & 40.7 & 30.4 & 36.3 & 22.2 & 38.8 & 36.9 & 35.3 \\ 
MIL+seg~\cite{Wsl} \raggedright & 79.6 & 50.2 & 21.6 & 40.9 & 34.9 & 40.5 & 45.9 & 51.5 & 60.6 & 12.6 & 51.2 & 11.6 & 56.8 & 52.9 & 44.8 & 42.7 & 31.2 & 55.4 & 21.5 & 38.8 & 36.9 & 42.0 \\ 
\hline
\textbf{Semi-supervised:} \raggedright & & & & & & & & & & & & & & & & & & & & & & \\
DecoupledNet~\cite{decouplednet} \raggedright & 86.5 & 69.9 & 33.6 & 58.5 & 42.4 & 50.4 & 68.8 & 63.2 & 67.5 & 11.5 & 61.8 & 20.0 & 61.2 & 66.7 & 60.1 & 50.8 & 30.2 & 67.9 & 33.9 & 59.2 & 51.0 & 53.1 \\
EM-Adapt~\cite{Wssl} \raggedright & - & - & - & - & - & - & - & - & - & - & - & - & - & - & - & - & - & - & - & - & - & 47.6 \\ 
\hline
\textbf{Transfer:} \raggedright & & & & & & & & & & & & & & & & & & & & & & \\
%TransferNet \raggedright & 85.0 & 66.8 & 24.7 & 60.3 & 40.7 & 46.8 & 69.4 & 52.2 & 70.0 & 11.2 & 70.2 & 19.3 & 63.6 & 66.6 & 62.4 & 41.3 & 28.6 & 72.1 & 33.9 & 52.7 & 37.6 & 51.2 \\ 
TransferNet \raggedright &85.3 &68.5 &26.4 &69.8 &36.7 &49.1 &68.4 &55.8 &77.3 &6.2 &75.2 &14.3 &69.8 &71.5 &61.1 &31.9 &25.5 &74.6 &33.8 &49.6 &43.7 & 52.1\\
TransferNet-GT \raggedright & 85.2 & 70.6 & 25.3 & 61.7 & 42.2 & 38.9 & 67.5 & 53.9 & 73.3 & 20.6 & 81.5 & 26.9 & 69.6 & 73.2 & 66.6 & 36.7 & 26.9 & 82.9 & 42.2 & 54.4 & 39.3 & 54.3\\ 
DecoupledNet$^\dagger$ \raggedright &79.2 &13.1 &7.7 &38.4 &14.3 &15.0 &14.7 &46.0 &60.5 &3.7 &28.0 &1.7 &54.0 &37.5 &24.0 &9.2 &4.5 &46.2 &3.4 &18.7 &13.0 &25.4 \\
BaselineNet \raggedright & 81.4 & 30.6 & 9.2 & 41.8 & 27.0 & 32.9 & 46.9 & 44.7 & 61.2 & 7.0 & 59.2 & 4.7 & 55.2 & 55.5 & 22.7 & 32.0 & 17.5 & 65.7 & 15.8 & 33.6 & 18.1 & 36.3 \\  
\hline
\end{tabular}
\label{tab:voc_result_val}
\end{table*}
%%%%%%%%%%%%%%%%%%%%%%%%%%%%%%%%%%%%%%%%%%
\begin{table*}[!t] \footnotesize
\centering
\caption{Evaluation results on PASCAL VOC 2012 \textit{test} set.} \vspace{0.1cm}
\begin{tabular}
{
@{}C{2.6cm}@{}|@{}C{0.68cm}@{}C{0.66cm}@{}C{0.66cm}@{}C{0.66cm}@{}C{0.66cm}@{}C{0.66cm}@{}C{0.66cm}@{}C{0.66cm}@{}C{0.66cm}@{}C{0.66cm}@{}C{0.66cm}@{}C{0.66cm}@{}C{0.66cm}@{}C{0.66cm}@{}C{0.75cm}@{}C{0.75cm}@{}C{0.66cm}@{}C{0.66cm}@{}C{0.66cm}@{}C{0.66cm}@{}C{0.66cm}@{}|@{}C{0.75cm}@{}
}
\hline
Method&bkg&aero&bike&bird&boat&bottle&bus&car&cat&chair&cow&table&dog&horse&mbk&person&plant&sheep&sofa&train&tv&mean\\
\hline
\textbf{Fully-supervised:} \raggedright & & & & & & & & & & & & & & & & & & & & & \\
FCN-8s~\cite{Fcn} \raggedright & 91.2 & 76.8 & 34.2 & 68.9 & 49.4 & 60.3 & 75.3 & 74.7 & 77.6 & 21.4 & 62.5 & 46.8 & 71.8 & 63.9 & 76.5 & 73.9 & 45.2 & 72.4 & 37.4 & 70.9 & 55.1 & 62.2 \\ 
% TTIC Zoomout~\cite{Zoomout} \raggedright & 89.8 & 81.9 & 35.1 & 78.2 & 57.4 & 56.5 & 80.5 & 74.0 & 79.8 & 22.4 & 69.6 & 53.7 & 74.0 & 76.0 & 76.6 & 68.8 & 44.3 & 70.2 & 40.2 & 68.9 & 55.3 & 64.4 \\
CRF-RNN~\cite{Crfrnn} \raggedright & 93.1 & 90.4 & 55.3 & 88.7 & 68.4 & 69.8 & 88.3 & 82.4 & 85.1 & 32.6 & 78.5 & 64.4 & 79.6 & 81.9 & 86.4 & 81.8 & 58.6 & 82.4 & 53.5 & 77.4 & 70.1 & 74.7\\ 
DeepLab-CRF~\cite{Deeplabcrf} \raggedright& 93.1 & 84.4 & 54.5 & 81.5 & 63.6 & 65.9 & 85.1 & 79.1 & 83.4 & 30.7 & 74.1 & 59.8 & 79.0 & 76.1 & 83.2 & 80.8 & 59.7 & 82.2 & 50.4 & 73.1 & 63.7 & 71.6 \\ 
DeconvNet~\cite{deconvnet} \raggedright & 93.1 & 89.9 & 39.3 & 79.7 & 63.9 & 68.2 & 87.4 & 81.2 & 86.1 & 28.5 & 77.0 & 62.0 & 79.0 & 80.3 & 83.6 & 80.2 & 58.8 & 83.4 & 54.3 & 80.7 & 65.0 & 72.5 \\
\hline
\textbf{Weakly-supervised:} \raggedright & & & & & & & & & & & & & & & & & & & & & \\
EM-Adapt~\cite{Wssl} \raggedright & 76.3 & 37.1 & 21.9 & 41.6 & 26.1 & 38.5 & 50.8 & 44.9 & 48.9 & 16.7 & 40.8 & 29.4 & 47.1 & 45.8 & 54.8 & 28.2 & 30.0 & 44.0 & 29.2 & 34.3 & 46.0 & 39.6 \\
CCNN~\cite{Ccnn} \raggedright & 70.1 & 24.2 & 19.9 & 26.3 & 18.6 & 38.1 & 51.7 & 42.9 & 48.2 & 15.6 & 37.2 & 18.3 & 43.0 & 38.2 & 52.2 & 40.0 & 33.8 & 36.0 & 21.6 & 33.4 & 38.3 & 35.6 \\
MIL+seg~\cite{Wsl} \raggedright & 78.7 & 48.0 & 21.2 & 31.1 & 28.4 & 35.1 & 51.4 & 55.5 & 52.8 & 7.8 & 56.2 & 19.9 & 53.8 & 50.3 & 40.0 & 38.6 & 27.8 & 51.8 & 24.7 & 33.3 & 46.3 & 40.6 \\
% CCNN w/ size~\cite{Ccnn} \raggedright & 79.5 & 42.3 & 24.5 & 56.0 & 30.6 & 39.0 & 58.8 & 52.7 & 54.8 & 14.6 & 48.4 & 34.2 & 52.7 & 46.9 & 61.1 & 44.8 & 37.4 & 48.8 & 30.6 & 47.7 & 41.7 & 45.1 \\
\hline
\textbf{Transfer:} \raggedright & & & & & & & & & & & & & & & & & & & & & \\
%TransferNet \raggedright &85.5 &70.8 &29.1 &66.7 &42.2 &42.9 &70.1 &52.1 &67.2 &11.7 &61.0 &21.2 &62.2 &67.4 &71.3 &42.7 &31.4 &69.1 &38.0 &43.0 &33.1 &51.4 \\
TransferNet \raggedright &85.7 &70.1 &27.8 &73.7 &37.3 &44.8 &71.4 &53.8 &73.0 &6.7 &62.9 &12.4 &68.4 &73.7 &65.9 &27.9 &23.5 &72.3 &38.9 &45.9 &39.2&51.2\\
\hline
\end{tabular}
\label{tab:voc_result_test}
\end{table*}
% \vspace{-0.1cm}
%%%%%%%%%%%%%%%%%%%%%%%%%%%%%%%%%%%%%%%%%%%%%

% \vspace{-0.1in}
\subsection{Implementation Details}
\label{sec:implementation_detail}

\vspace{-0.05in}
\paragraph{Datasets}
We employ PASCAL VOC 2012~\cite{Pascalvoc} as target domain and Microsoft COCO (MS-COCO)~\cite{Mscoco} as source domain, which have 20 and 80 labeled semantic categories, respectively.
To simulate the transfer learning scenario, we remove all training images containing 20 PASCAL categories from MS-COCO dataset, and use only 17,443 images from 60 categories (with no overlap with the PASCAL categories) to construct the source domain data.
%To simulate the transfer learning scenario, we removed all images containing 20 PASCAL categories from MS-COCO dataset.
%Final training set for MS-COCO dataset is consisted of 17,443 images from 60 categories.
% We train our model using image-level class labels and segmentation annotations in PASCAL VOC and MS-COCO datasets, respectively, and evaluate the performance on PASCAL VOC 2012 benchmark images.
We train our model using image-level class labels in PASCAL VOC dataset and segmentation annotations in MS-COCO dataset, respectively, and evaluate the performance on PASCAL VOC 2012 benchmark images.
%Since there is neither segmentation annotations nor additional weakly annotations for PASCAL categories, our algorithm is directly comparable to other weakly-supervised approaches.

\iffalse
We evaluate our algorithm with PASCAL VOC~\cite{Pascalvoc} datasets.
It contains 20 semantic categories and ?? training images.
We employed MSCOCO~\cite{Mscoco} dataset as images with segmentation annotations. 
Originally the dataset contains ?? training images from 80 semantic categories, which can be considered as superset of PASCAL VOC dataset as all 20 PASCAL categories are exactly overlapped.
To simulate transfer learning scenario, we removed all images from MSCOCO dataset containing object from PASCAL categories.
Final training set for MSCOCO dataset is consisted of ?? images from 60 categories with segmentation annotation.
Note that there is no direct supervision for all PASCAL categories since we use segmentation annotations from categories exclusive to PASCAL dataset, so training dataset for  
20 PASCAL categories is identical to other weakly-supervised approaches.
\fi

\vspace{-0.1in}
\paragraph{Training}
We initialize the encoder by fine-tuning the pre-trained CNN from ImageNet~\cite{Imagenet} to perform multi-class classification on combined datasets of PASCAL VOC and MS-COCO.
The weights in the attention model and classification layers ($\theta_\alpha$ and $\theta_c$, respectively) are pre-trained by optimizing Eq.~\eqref{eqn:obj_cls}.
Then we optimize both decoder, attention model and classification layers jointly using the objective function in Eq.~\eqref{eqn:obj_all}, while the weights in the decoder ($\theta_s$) are initialized with zero-mean Gaussians.
We fix the weights in the encoder ($\theta_e$) during training.

\iffalse
We use the VGG-16 layer net as our encoder architecture, where the weights are initialized by the pre-trained model with Imagenet~\cite{Imagenet} database.
%1. pretrain on ImageNet~\cite{}.
2. finetune on combined MSCOCO+VOC for classification
3. fixing encoder (unitl the last convolutional layer), train attention model in~\ref{sec:attention}.
4. given attention, jointly train both attention and decoder.
\fi

\vspace{-0.1in}
\paragraph{Optimization}
We implement the proposed algorithm based on Caffe~\cite{Caffe} library. 
We employ Adam optimization~\cite{Adamsolver} to train our network with learning rate 0.0005 and default hyper-parameter values proposed in \cite{Adamsolver}.
The size of mini-batch is set to 64.
%For each mini-batch, we alternatively optimize Eq.~\eqref{} and \eqref{}, and the size of mini-batch is set to 64.
%The stochastic gradient descent with adaptive momentum~\cite{Adamsolver} is employed for optimization.
% Training our model takes 4 and 10 hours using Nvidia Titan X GPU for the pre-training of attention model including classification layers and the joint training of all other parts, respectively.
We trained our models using a NVIDIA Titan X GPU. Training our model takes 4 hours for the pre-training of attention model including classification layers, and 10 hours for the joint training of all other parts, respectively.

\iffalse
Solver parameters.
Training time.
\fi

\vspace{-0.1in}
\paragraph{Inference}
We exploit an additional classifier trained on PASCAL VOC dataset to predict class labels on target domain images.
The predicted class labels are used to generate segmentations as described in Section~\ref{sec:trainandinf}.
Optionally, we employ post processing based on fully-connected CRF~\cite{Fullycrf}.
In this case, we apply the CRF on foreground/background probability maps for each class label independently, and obtain combined segmentations by taking pixel-wise maximums of foreground probabilities across labels. 
%Given predicted labels, we obtain the segmentation label  segmentation label  which is then used to generate labels as described in \ref{sec:trainandinf}.

%%%%%%%%%%%%%%%%%%%%%%%%%%%%%%%%%%%%%%%%%%%%%
%\vspace{-0.05in}
\subsection{Comparison to Other Methods}
\label{sec:compare_to_other_method}
%\vspace{-0.05in}

\ifdefined\paratitle {\color{blue} 
[Descriptions for evaluation metric: mean IoU] \\
} \fi
This section presents evaluation results of our algorithm with the competitors on PASCAL VOC 2012 benchmark dataset.
We follow {\em comp6} evaluation protocol, and scores are measured by computing Intersection over Union (IoU) between ground truth and predicted segmentations.

\ifdefined\paratitle {\color{blue} 
[Comparison to weakly-supervised semantic segmentation algorithms.] \\
} \fi
Table~\ref{tab:voc_result_val} summarizes the evaluation results on PASCAL VOC 2012 validation dataset.
We compared the proposed algorithm with state-of-the-art weakly- and semi-supervised algorithms\footnote{Strictly speaking, our method is not directly comparable to both approaches since we use auxiliary examples.
% Note that there is no direct supervision for the categories in the target domain, and
Note that we do not use ground-truth segmentation annotations for the categories used in evaluation, since the examples are from different categories.
}.
\iffalse
{\color{blue}
%To understand impact of using annotations from different categories, we compared our method with weakly- and semi-supervised approaches.
We compared our method with weakly- and semi-supervised approaches. %to understand benefits of using annotations from different categories in weakly-supervised setting.
Strictly speaking, our method is not direct comparable to both approaches, since we are proposing the new problem setting, which is different to both settings. 
However, we argue that employing additional annotations from different categories is not a definite advantage over these settings as we do not exploit ground-truth annotations and other weakly- and semi-supervised approaches are not able to exploit the data. 
}
\fi
Our method is denoted by TransferNet, and TransferNet-GT indicates our method with ground-truth class labels for segmentation inference, which serves as the upper-bound performance of our method since it assumes classification is perfect.
%It shows the performance of our method when the classification is perfect, which is beneficial to isolate shape generation performance.
% The proposed algorithm outperforms all weakly-supervised semantic segmentation techniques~\cite{Wsl,Wssl,Ccnn} with substantial margins, although it does not employ any ground-truth segmentations for categories used in evaluation.
The proposed algorithm outperforms all weakly-supervised semantic segmentation techniques with substantial margins, although it does not employ any ground-truth segmentations for categories used in evaluation.
% The performance of the proposed algorithm is comparable to semi-supervised semantic segmentation methods~\cite{decouplednet, Wssl}, which exploits a small number of ground-truth segmentations in addition to weakly-annotated images for training.
The performance of the proposed algorithm is comparable to semi-supervised semantic segmentation methods, which exploits a small number of ground-truth segmentations in addition to weakly-annotated images for training.
The results suggest that segmentation annotations from different categories can be used to make up the missing supervision in weakly-annotated images; the proposed encoder-decoder architecture based on attention model successfully captures transferable segmentation knowledge from the exclusive segmentation annotations and uses it as prior for segmentation in unseen categories. 

Table~\ref{tab:voc_result_test} summarizes our results on PASCAL VOC 2012 test dataset.
% Our algorithm exhibits superior performance to weakly-supervised approaches~\cite{Wsl,Wssl,Ccnn}, but there are still large performance gaps with respect to fully-supervised approaches~\cite{Fcn,deconvnet,Zoomout,Deeplabcrf}.
Our algorithm exhibits superior performance to weakly-supervised approaches, but there are still large performance gaps with respect to fully-supervised approaches.
% This is because some target categories exhibits unique shapes, which is difficult to capture with the model trained in source domain.  
It shows that there is domain-specific segmentation knowledge which cannot be made up by annotations form different categories. 
% and reducing this gap would be interesting future work.
% Reducing this gap would be interesting future work.

%{\color{red} It shows the gap between general and domain-specific segmentation knowledge, and reducing the gap would be another interesting future work.}

\begin{figure*}[!t]
\hspace{.8cm}Input Image ~\hspace{.7cm} Ground-truth \hspace{.6cm} Densified attention \hspace{.55cm} BaselineNet \hspace{1.1cm}TransferNet \hspace{.6cm} TransferNet+CRF\\
\vspace{-0.2cm}
%\includegraphics[width=0.98\linewidth] {../figure/Overall_Results/legend.pdf}\\
%\begin{center}
\subfigure[Examples that our method produces accurate segmentation.]{
\begin{minipage}{1\textwidth}
\centering
\includegraphics[width=0.161\linewidth] {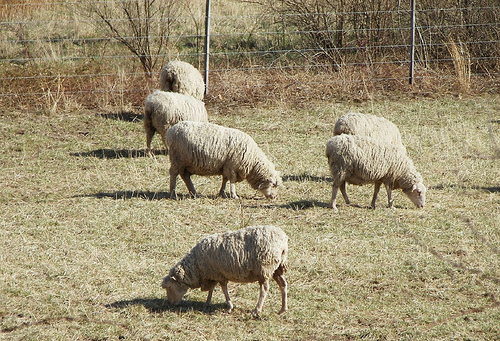}
\includegraphics[width=0.161\linewidth] {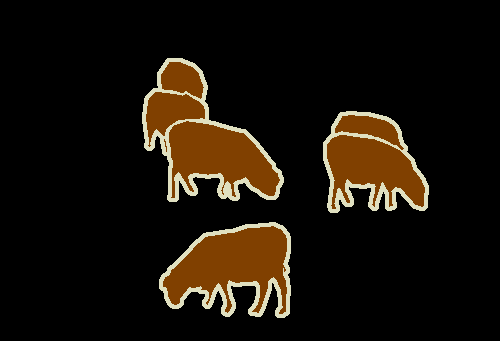}
\includegraphics[width=0.161\linewidth] {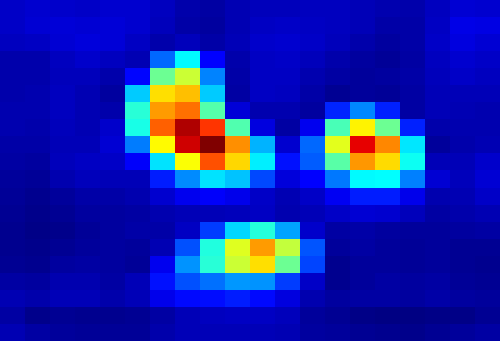}
\includegraphics[width=0.161\linewidth] {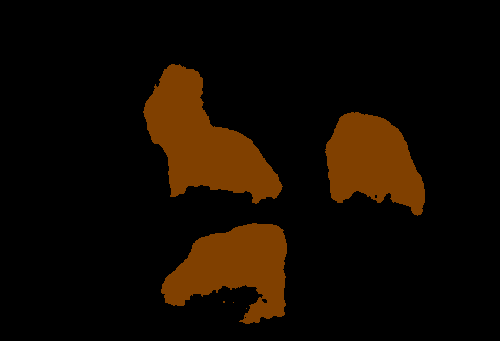}
\includegraphics[width=0.161\linewidth] {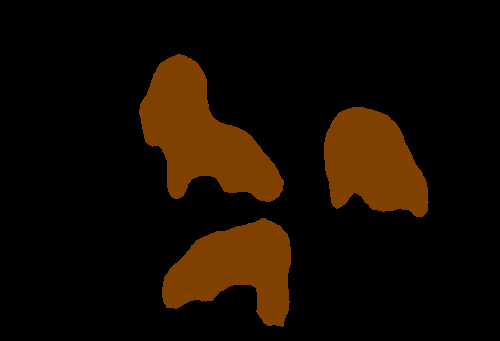}
\includegraphics[width=0.161\linewidth] {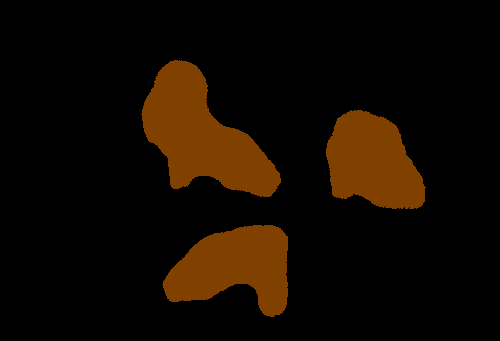}\\
\includegraphics[width=0.161\linewidth] {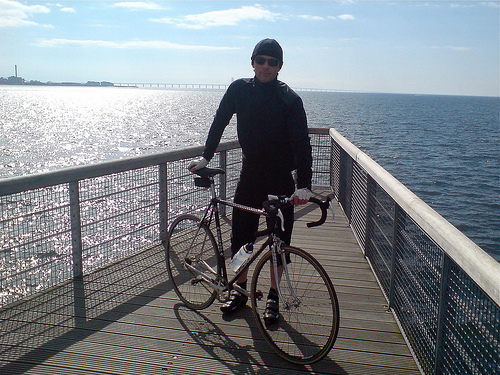}
\includegraphics[width=0.161\linewidth] {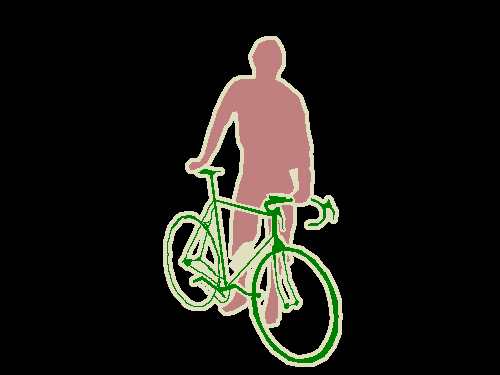}
\includegraphics[width=0.161\linewidth] {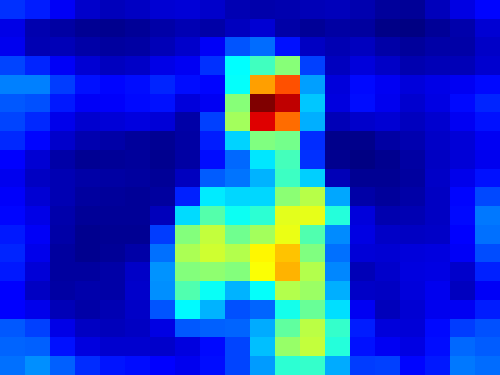}
\includegraphics[width=0.161\linewidth] {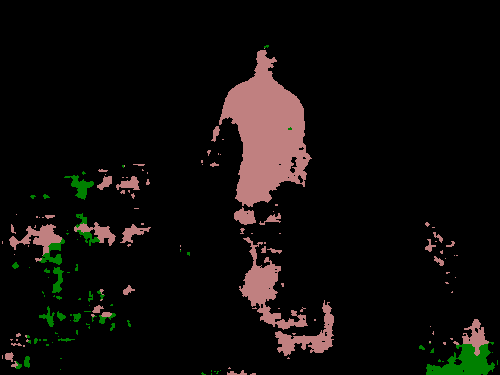}
\includegraphics[width=0.161\linewidth] {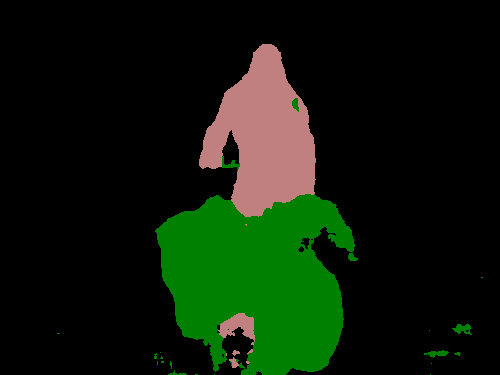}
\includegraphics[width=0.161\linewidth] {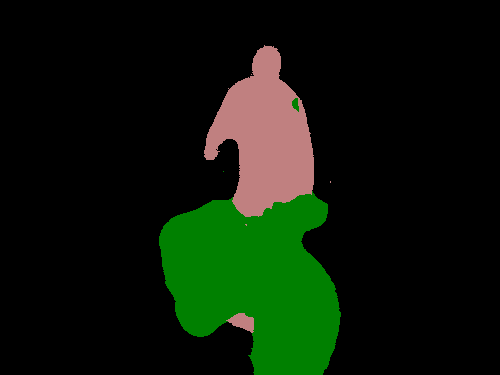}\\
\includegraphics[width=0.161\linewidth] {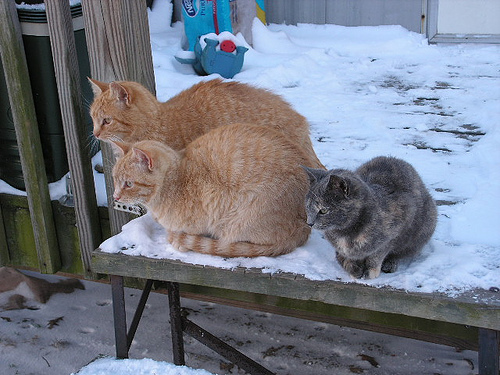}
\includegraphics[width=0.161\linewidth] {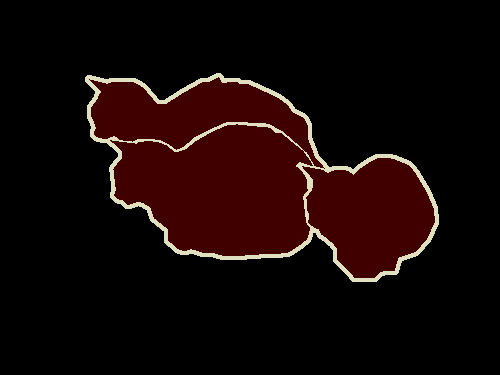}
\includegraphics[width=0.161\linewidth] {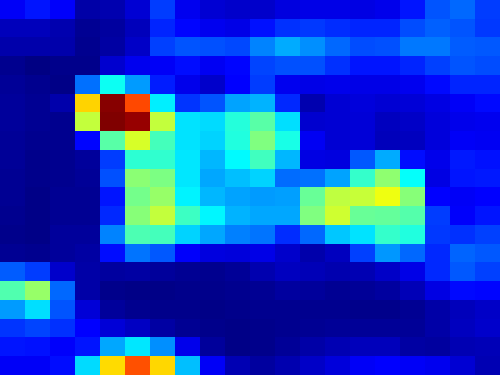}
\includegraphics[width=0.161\linewidth] {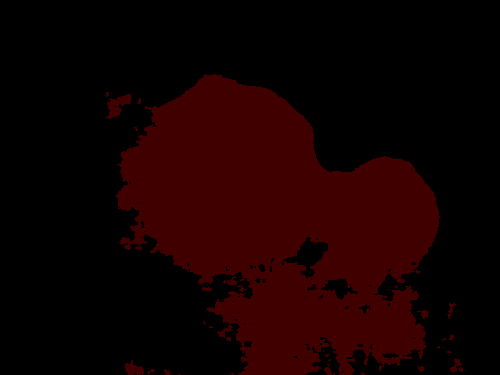}
\includegraphics[width=0.161\linewidth] {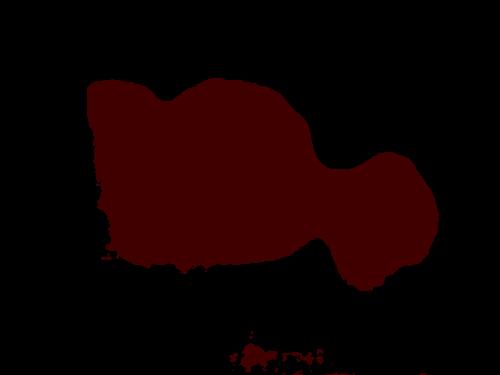}
\includegraphics[width=0.161\linewidth] {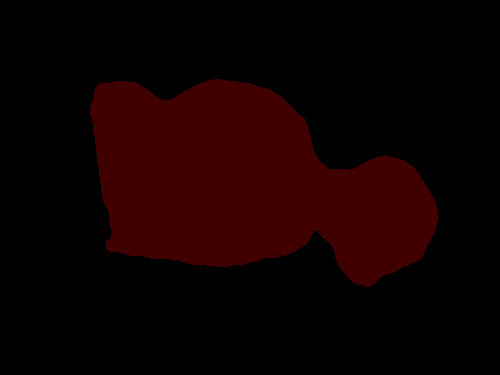}\\
\includegraphics[width=0.161\linewidth] {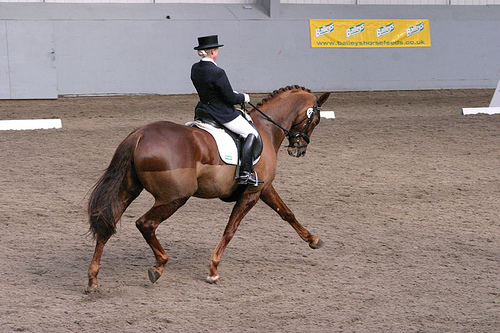}
\includegraphics[width=0.161\linewidth] {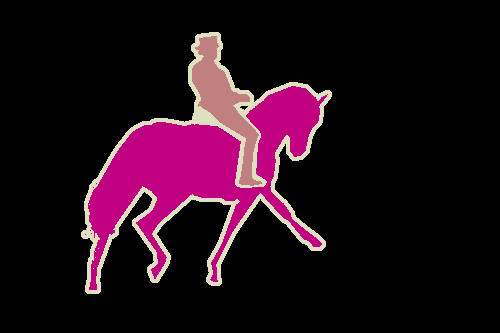}
\includegraphics[width=0.161\linewidth] {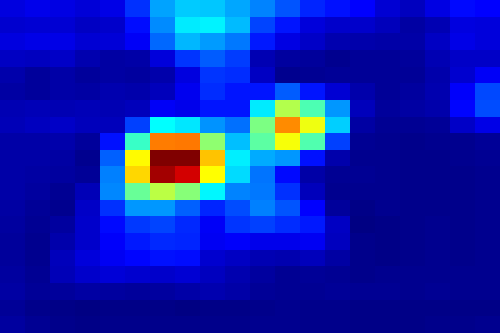}
\includegraphics[width=0.161\linewidth] {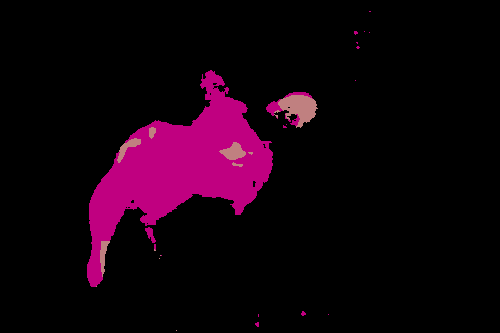}
\includegraphics[width=0.161\linewidth] {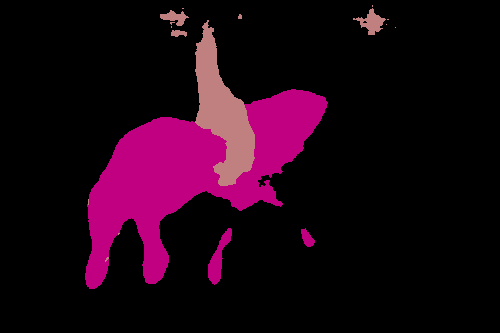}
\includegraphics[width=0.161\linewidth] {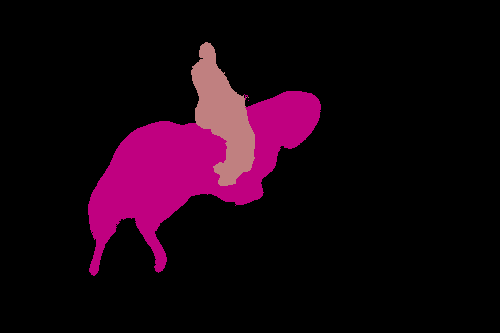}\\
\includegraphics[width=0.161\linewidth] {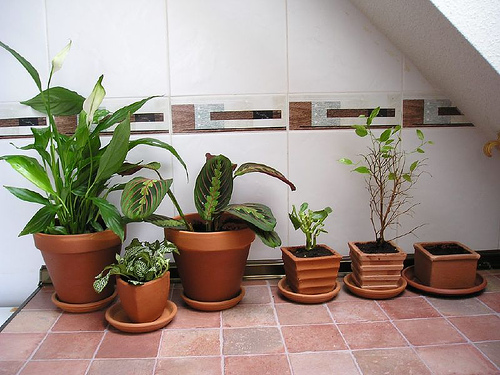}
\includegraphics[width=0.161\linewidth] {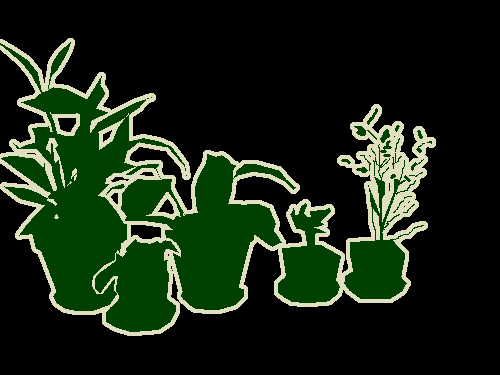}
\includegraphics[width=0.161\linewidth] {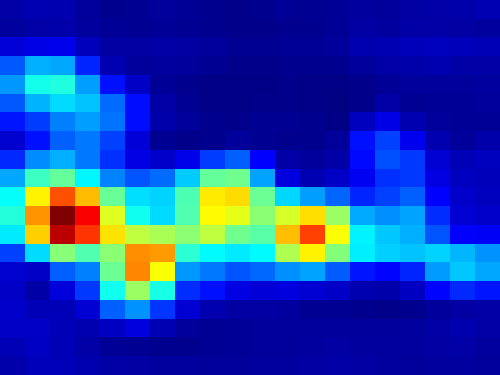}
\includegraphics[width=0.161\linewidth] {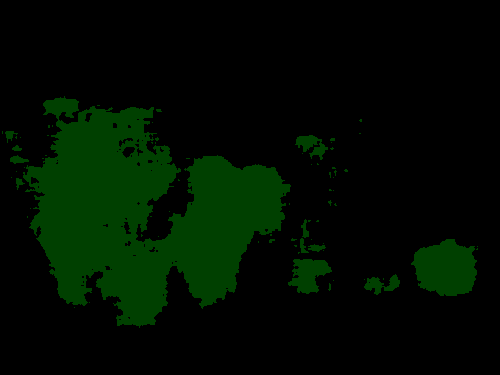}
\includegraphics[width=0.161\linewidth] {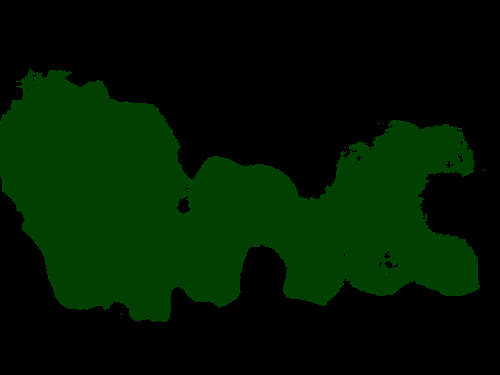}
\includegraphics[width=0.161\linewidth] {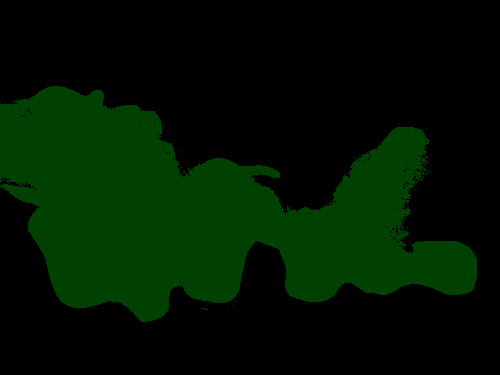}\\
\includegraphics[width=0.161\linewidth] {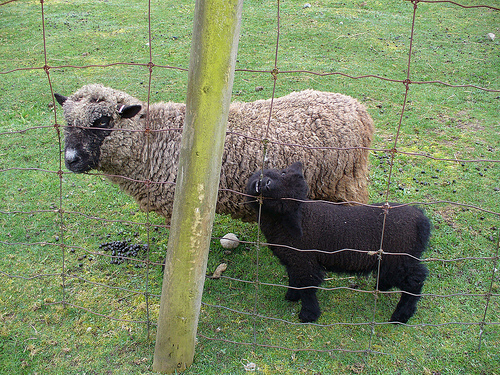}
\includegraphics[width=0.161\linewidth] {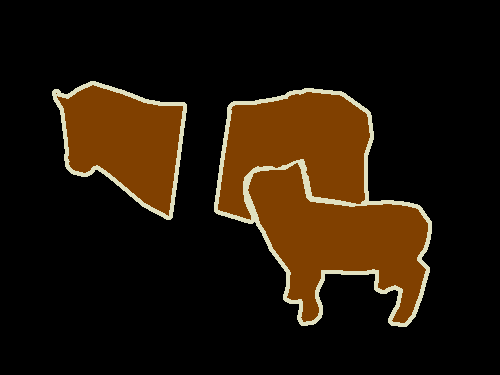}
\includegraphics[width=0.161\linewidth] {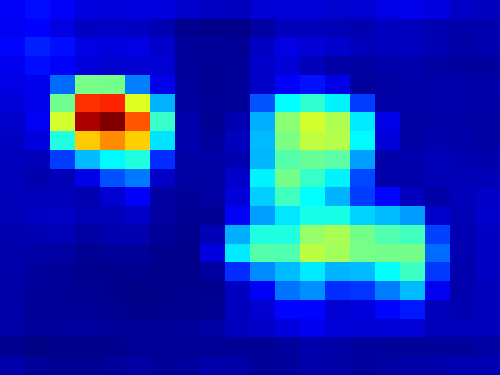}
\includegraphics[width=0.161\linewidth] {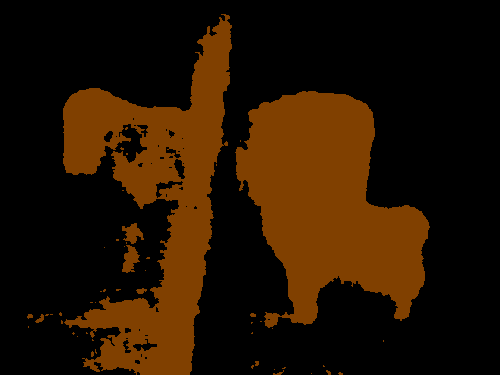}
\includegraphics[width=0.161\linewidth] {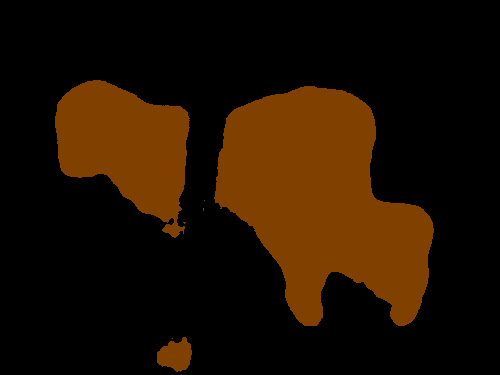}
\includegraphics[width=0.161\linewidth] {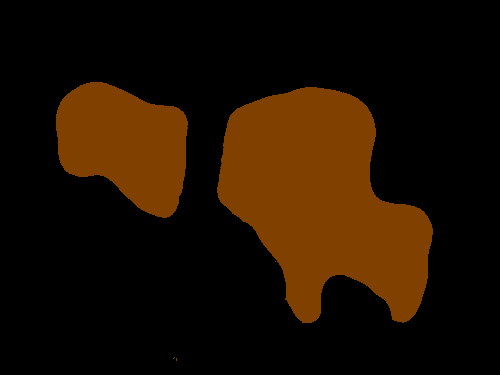}\\
\includegraphics[width=0.161\linewidth] {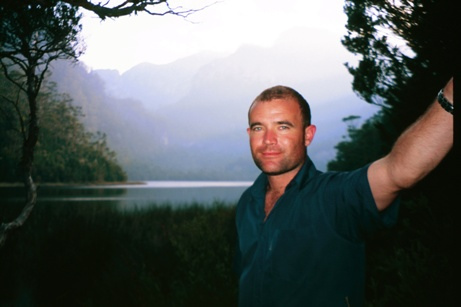}
\includegraphics[width=0.161\linewidth] {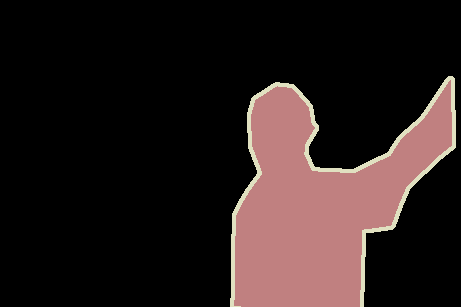}
\includegraphics[width=0.161\linewidth] {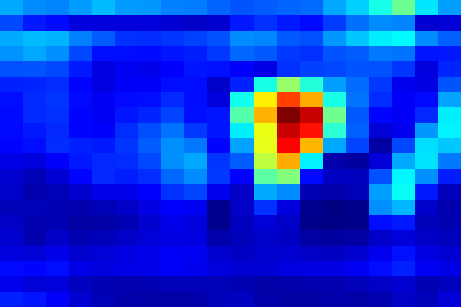}
\includegraphics[width=0.161\linewidth] {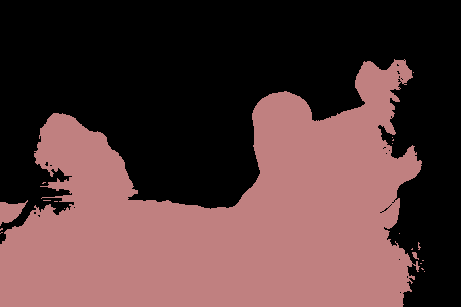}
\includegraphics[width=0.161\linewidth] {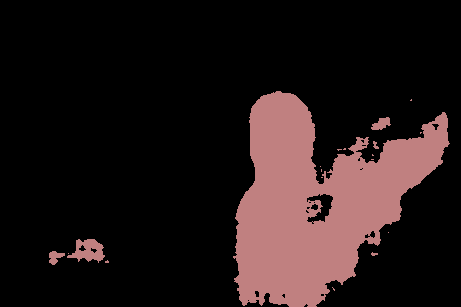}
\includegraphics[width=0.161\linewidth] {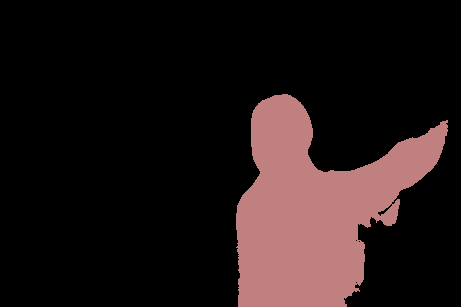}\\
\vspace{0,1cm}
\end{minipage}
}
\subfigure[Examples that our method produces inaccurate segmentation due to misclassification (top) or inaccurate attention (bottom).]{
\begin{minipage}{1\textwidth}
\centering
\includegraphics[width=0.161\linewidth] {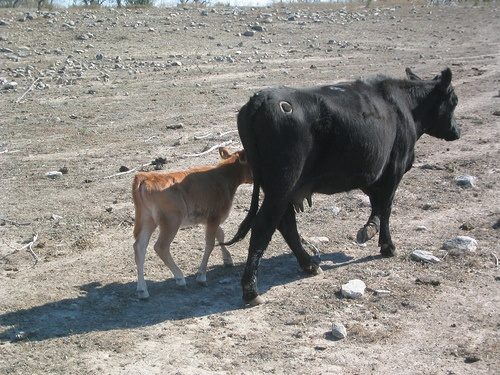}
\includegraphics[width=0.161\linewidth] {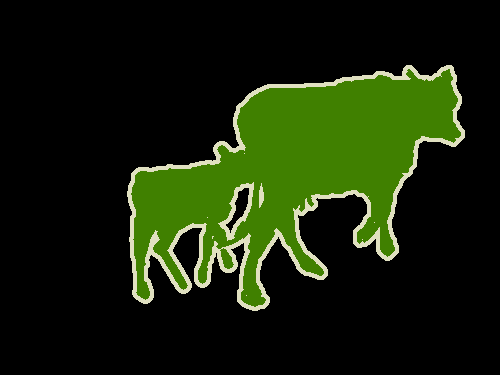}
\includegraphics[width=0.161\linewidth] {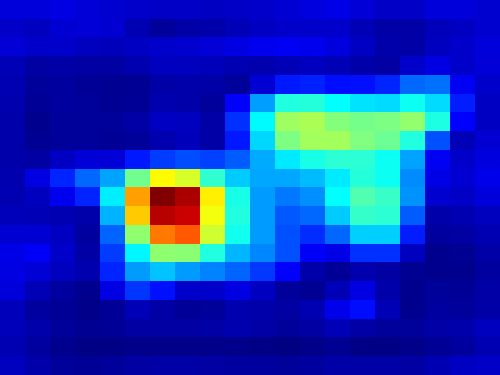}
\includegraphics[width=0.161\linewidth] {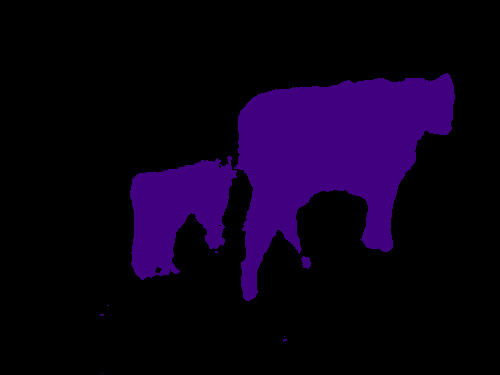}
\includegraphics[width=0.161\linewidth] {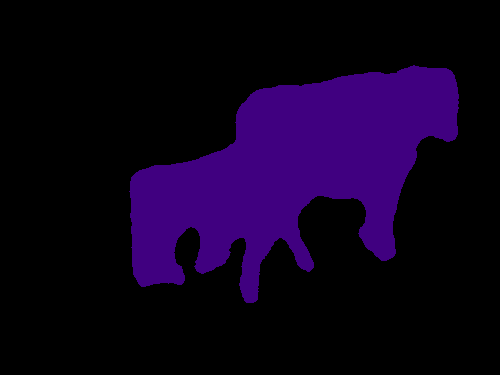}
\includegraphics[width=0.161\linewidth] {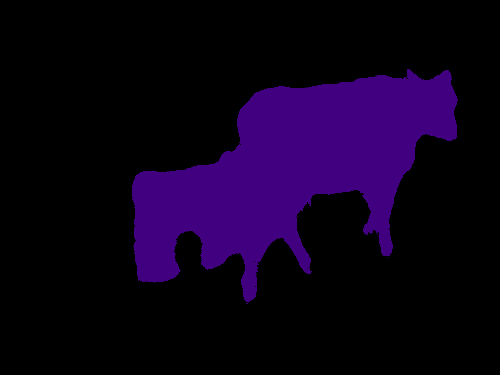}\\
\includegraphics[width=0.161\linewidth] {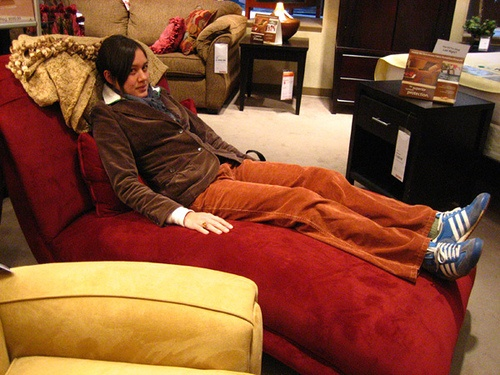}
\includegraphics[width=0.161\linewidth] {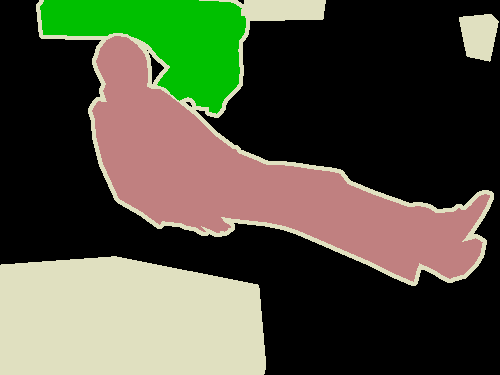}
\includegraphics[width=0.161\linewidth] {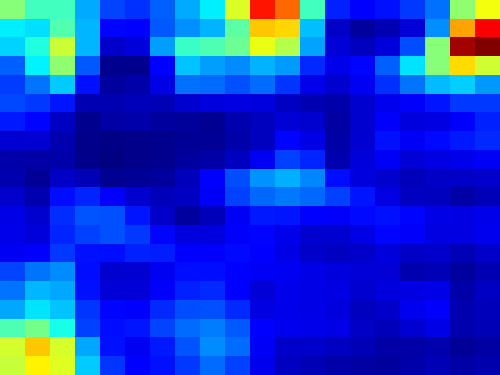}
\includegraphics[width=0.161\linewidth] {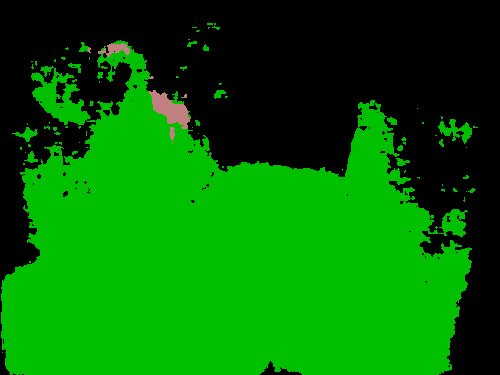}
\includegraphics[width=0.161\linewidth] {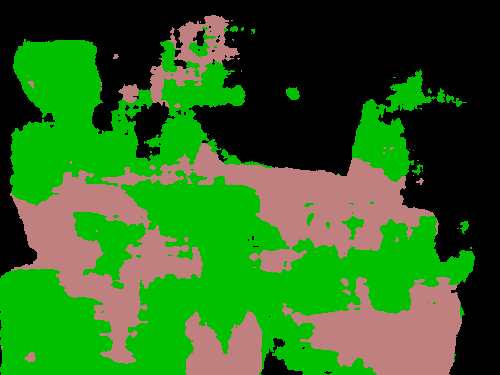}
\includegraphics[width=0.161\linewidth] {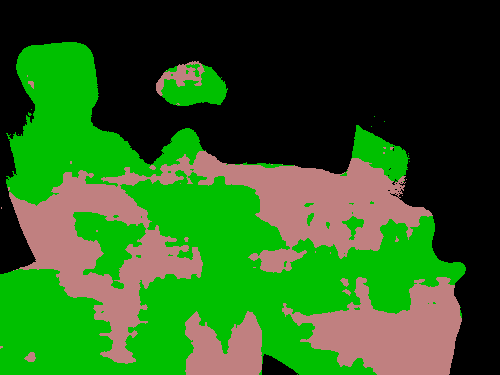}\\
\vspace{0.1cm}
\end{minipage}
\label{fig:qualitative_result_failure}
}
%\end{center}
\caption{
Examples of semantic segmentation on PASCAL VOC 2012 validation images. 
The attentions (3rd column) are extracted from the model trained using Eq.~\eqref{eqn:obj_all}, and aggregated over all categories for visualization. 
(a) Our methods based on attention model (TransferNet and TransferNet+CRF) produce accurate segmentation result even without CRF by transferring learned segmentation knowledge from source domain.
Our results tend to be denser and more accurate than the results from BaselineNet, which generates segmentation from class score map.
(b) Our algorithm may produce inaccurate segmentation when the input labels are wrong due to misclassification (top) or attention output is noisy (bottom). 
}
\label{fig:qualitative_result}
\end{figure*}

The qualitative results of the proposed algorithm are presented in Figure~\ref{fig:qualitative_result}.
Our algorithm often produces accurate segmentations in the target domain by transferring the decoder trained with source domain examples, although it is not successful in capturing some category-specific fine details in some examples.
The missing details can be recovered through post-processing based on CRF.
Since the attention model in the target domain may not be perfect due to missing supervisions, our algorithm sometimes produces noisy predictions as illustrated in Figure~\ref{fig:qualitative_result_failure}.
See Apendix~\ref{apdx_sec:additional_results} for more qualitative comparisons.

%\ref{fig:qualitative_result}. 
%It is mainly because of domain-specific knowledge which cannot be captured by 
%Compared to semi-supervised semantic segmentation algorithms which exploits small number of segmentation annotations in addition to weakly-annotated images, the performance of our algorithm is almost   
\iffalse
Our method significantly outperforms existing the state-of-the-art weakly-supervised semantic segmentation algorithms with a substantial margin, and even compatible to semi-supervised algorithms, although we do not use any ground-truth segmentation for the target categories.
It shows that there are some general segmentation kwoledge that can be benefitial for segmentations in general categories, and our algorithm captures that information effectively. 
\fi

\ifdefined\paratitle {\color{blue} 
[Benefit of our architecture over other baselines] \\
} \fi
\subsection{Comparison to Baselines}
\label{sec:compare_to_other_baselines}
To better understand the benefits from the attention model in our transfer learning scenario, we compare the proposed algorithm with two baseline algorithms, which are denoted by DecoupledNet$^\dagger$ and BaselineNet.
(See Appendix \ref{apdx_sec:archs} for detailed configurations of the baselines and proposed algorithm.)
%\footnote{Detailed descriptions and configurations of the baseline algorithms are presented in the supplementary file.}.

DecoupledNet$^\dagger$ directly applies the architecture proposed in \cite{decouplednet} to our transfer learning task by training the decoder in MS-COCO and applying it to PASCAL VOC for inference. 
The model employs the same decoupled decoder architecture to ours, but has a direct connection between encoder and decoder without attention mechanism.
%The result in table~\ref{tab:voc_result_val} shows that the model is not generalizable between different domains.
The result in Table~\ref{tab:voc_result_val} shows that the model trained on source domain fails to adapt to target domain categories.
%It is mainly because the representation from the encoder changes in different domains, and the decoder is not working for unseen representations.
It is mainly because the decoder cannot interpret the features from unseen categories in target domain.
Our model can mitigate the issue since the attention model provides coherent representations to decoder across domains.

Although the above baseline shows the benefits of the attention model in our architecture, the advantage of attention estimation from the intermediate layer may be still ambiguous.
To this end, we employ another FCN~\cite{Fcn}-style baseline denoted by BaselineNet, which uses class score map as input to the decoder.
% To this end, we employ another baseline denoted by BaselineNet, which uses class score map as input to the decoder similar to FCN~\cite{Fcn}.
%BaselineNet is another baseline that uses class score map as input to the decoder similar to FCN~\cite{Fcn}.
It can be considered as a specific case of our method that the attention is extracted from the final layer of the classification network ($f_{\text{att}}=f_{\text{cls}}$).
%The performance of BaselineNet is better than DecoupledNet$^\dagger$ since the class score map is more general representation to decoder.
The performance of BaselineNet is better than DecoupledNet$^\dagger$ since the class score map is a general representation to the decoder across different categories.
However, the performance is considerably worse than the proposed method as shown in Table~\ref{tab:voc_result_val} and Figure~\ref{fig:qualitative_result}.
We observe that the class score map is sparse and in row-resolution, while densified attention map in our model contains richer information for segmentation. 

The comparisons to the baseline algorithms show that transferring segmentation knowledge from different categories is a very challenging task.
The straightforward extensions of existing architectures have difficulty in generalizing the knowledge across different categories.
In contrast, our model effectively transfers segmentation knowledge by learning general features through attention mechanism.

\begin{figure}[!t]
\centering
\includegraphics[width=0.97\linewidth] {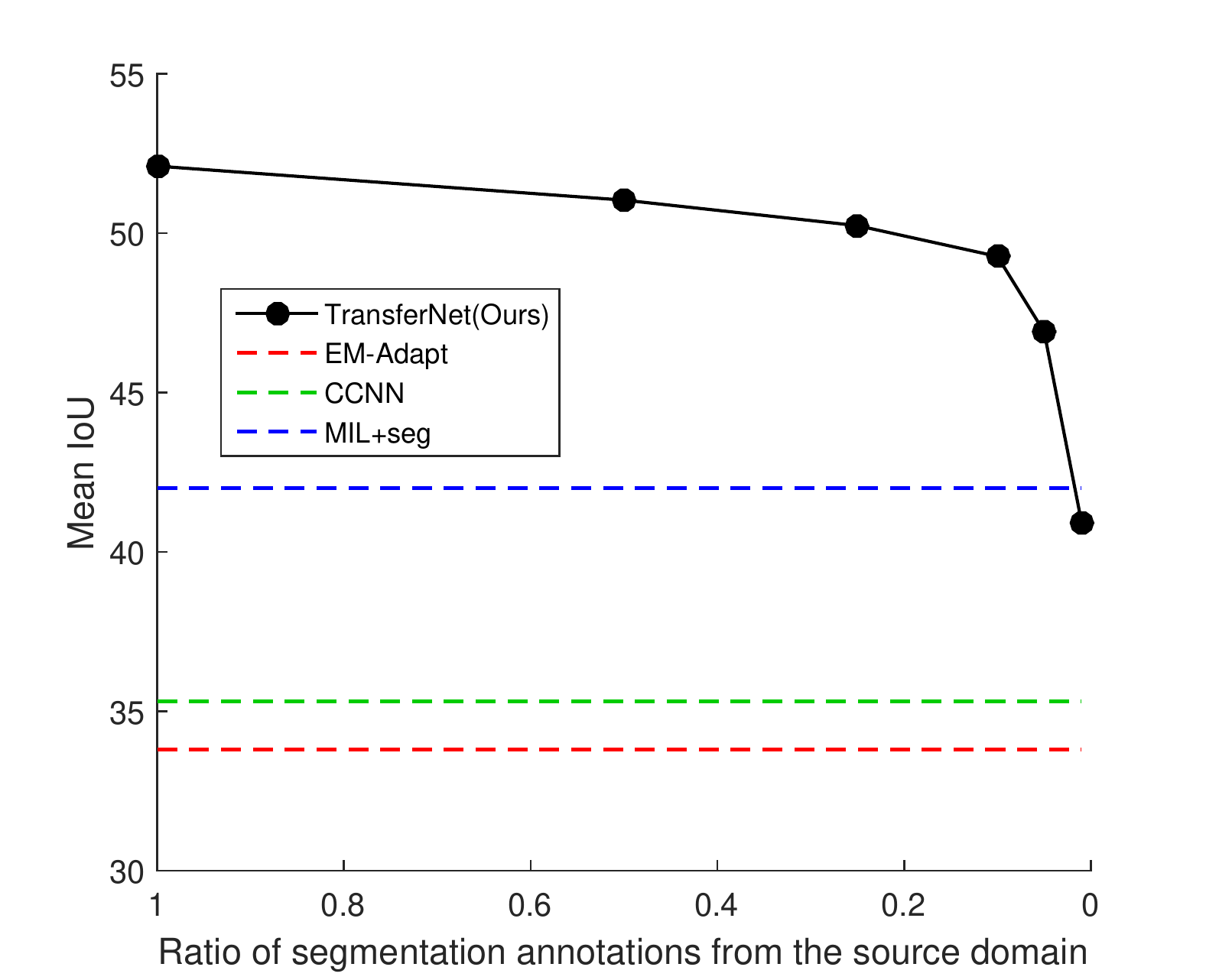} 
\caption{ Performance of the proposed algorithm with varying number of annotations in the source domain. }
\label{fig:ctrl_numanno}
\end{figure}

\ifdefined\paratitle {\color{blue} 
[Effect of number of strong annotations in the source domain] \\
} \fi
\subsection{Impact of Number of Annotations in the Source Domain}
\label{sec:effect_of_anno}
To see the impact of number of annotations in the source domain, we conduct additional experiments by varying the number of annotations in the source domain (MS-COCO).
To this end, we randomly construct subsets of training data by varying their sizes in ratios (50\%, 25\%, 10\%, 5\% and 1\%) and average the performance in each size with 3 subsets.
The results are illustrated in Figure~\ref{fig:ctrl_numanno}.
In general, more annotations in the source domain improve the segmentation quality on the target domain.
Interestingly, the performance of the proposed algorithm is still better than other weakly-supervised methods even with a very small fraction of annotations.
It suggests that exploiting even small number of segmentations from other categories can effectively reduce the gap between the approaches based on strong and weak supervisions.

\section{Conclusion}
\label{sec:conclusion}
%We propose a novel weakly-supervised semantic segmentation algorithm based on deep neural network, which aims to make up the missing supervisions on segmentation by transferring segmentation knowledge using strong annotations from other categories.
We propose a novel approach for weakly-supervised semantic segmentation, which exploits extra segmentation annotations in different categories to improve segmentation on the dataset with missing supervisions.
The proposed encoder-decoder architecture with attention model is appropriate to capture transferable segmentation knowledge across categories. 
The results on challenging benchmark dataset suggest that the gap between approaches based on strong and weak supervision can be reduced by transfer learning.
We believe that scaling up the proposed algorithm to a large number of categories would be one of interesting future research direction (\eg, semantic segmentation on 7.6K categories in ImageNet dataset using segmentation annotations from 20 PASCAL VOC categories).
%The proposed algorithm is potentially very useful to scale up the semantic segmentation problem to a large number of categories (e,g semantic segmentation on 7.6K categories in Imagenet dataset using segmentation annotations from 20 PASCAL VOC categories), which can be interesting future research direction.

\clearpage
{\small
\bibliographystyle{ieee}
\bibliography{egbib}
}
\clearpage
% ====================================================================
% Appendix
% !TEX root = cvpr2016_transfer.tex
\appendix
\section*{Appendix}
\addcontentsline{toc}{section}{Appendix}
\renewcommand{\thesection}{\Alph{section}}

\section{Analysis of Densified Attention}
\label{apdx_sec:densified_att}

This section describes more comprehensive analysis of the densified attention discussed in Section~\ref{sec:segmentation}. 
By plugging Eq.~\eqref{eqn:context_vector} into Eq.~\eqref{eqn:decoder_input}, we obtain the densified attention, which is given by
\begin{equation}
{\bf s}^l = {\bf A}{\bf A}^\mathrm{T}{\boldsymbol \alpha}^l = {\bf G}{\boldsymbol \alpha}^l, 
\label{apdx_eqn:densified_att}
\end{equation}
where ${\bf G} \in \mathrm{R}^{M \times M}$ is the Gram matrix of ${\bf A}\in\mathrm{R}^{M\times D}$; each element in  ${\bf G}$ denoted by $G_{ij} \equiv \sum_k A_{ik} A_{jk}$ represents similarity between $i^{\text{th}}$ and $j^{\text{th}}$ pixels in the feature map.
%Based on Eq.~\eqref{apdx_eqn:densified_att}, the densified attention is given by the weighted linear combination of rows in the Gram matrix which represents  similarity of each pixel to all other regions in the image, and the weights are given by the attention ${\boldsymbol \alpha}^l$.
Eq.~\eqref{apdx_eqn:densified_att} means that the densified attention is given by the weighted linear combination of rows in the Gram matrix, where the weights are given by the attention ${\boldsymbol \alpha}^l\in \mathrm{R}^M$.
The densified attention ${\bf s}^l$ reveals more detailed shape of the objects than the sparse attention weight ${\boldsymbol \alpha}^l$ by highlighting not only attended pixels but also visually correlated areas.
%Compared to the sparse attention weight ${\boldsymbol \alpha}^l \in \mathrm{R}^M$, the densified attention ${\bf s}^l$ reveals more dense and detailed shape of the objects by highlighting not only pixels focused by attention model but also other regions similar to the focused regions.
%Since the Gram matrix is category-independent representation, the importance of each row is controlled by category-specific attention ${\boldsymbol \alpha}^l$, and generates
% and each row in the Gram matrix represents similarity of each pixel to all other regions in the image
\iffalse
where $\bf{G}\in\mathrm{R}^{M\times M}$ is the Gram matrix of the feature map $\bf{A}\in\mathrm{R}^{M\times D}$.
Each element of Gram matrix $G_{ij} = \sum_k A_{ik} A_{jk}$ represents similarity between spatial units in the feature map.
Compared to the sparse attention weight ${\boldsymbol \alpha}^l \in \mathrm{R}^M$, the densified attention ${\bf s}^l$ reconstructs more dense and detailed shape of the objects by highlighting not only pixels focused by attention model but also other regions similar to the focused regions.
\fi

\begin{figure}[!t]
\centering
\includegraphics[width=1\linewidth] {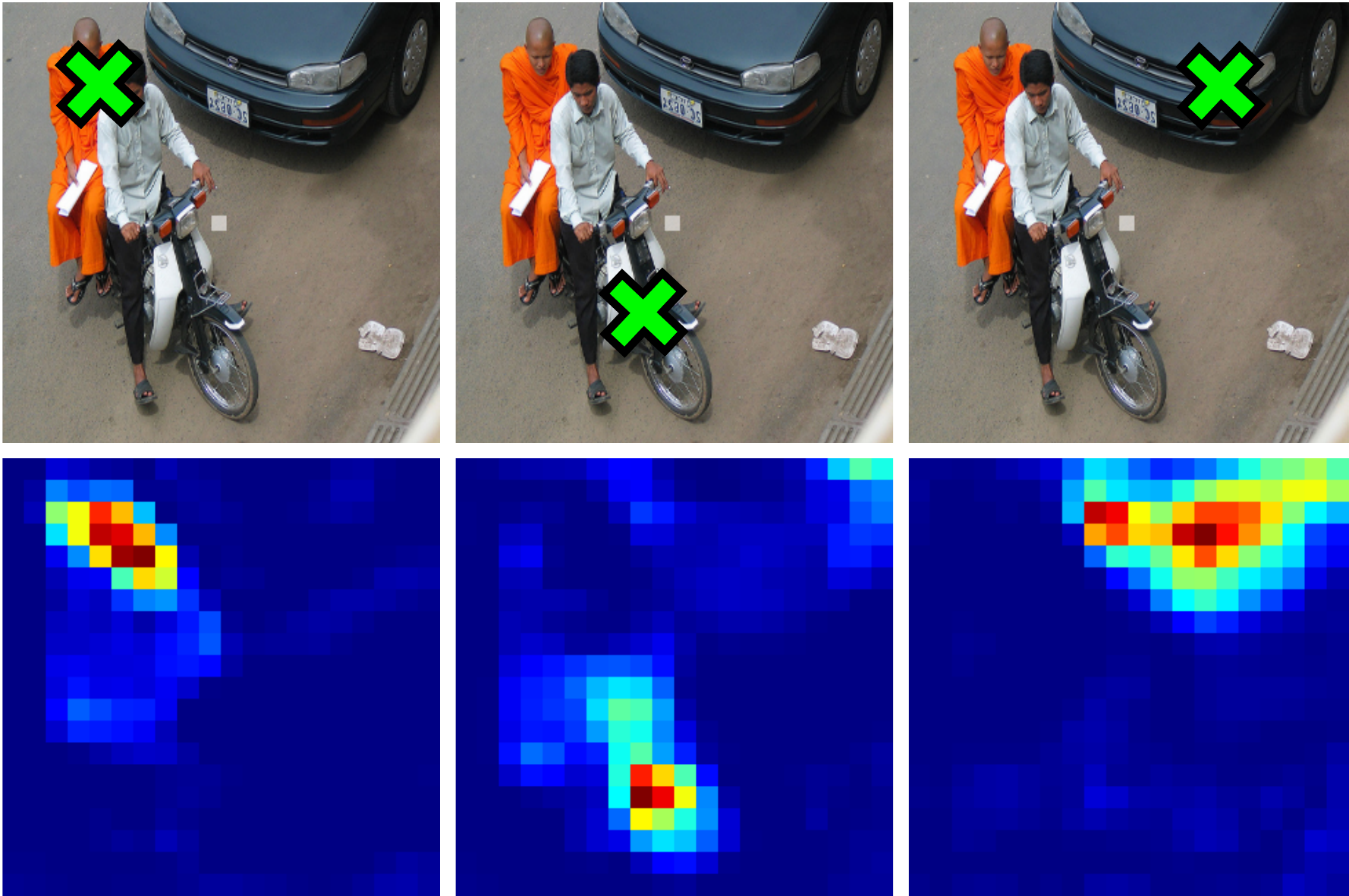}
\caption{
Visualizations of selected rows in the Gram matrix. (Top row) input image. (Bottom row) visualizations of rows in the Gram matrix which are selected from the pixel location marked on the image. Each row is reshaped to original spatial shape of the feature map for effective visualization. }
\label{fig:apdx_vis_gram}
\end{figure}

Figure~\ref{fig:apdx_vis_gram} visualizes the densified attention obtained by selecting a row of the Gram matrix in Eq.~\eqref{apdx_eqn:densified_att}.
The row is given by the one-hot attention vector ${\boldsymbol \alpha}^l$, which represents a single pixel attention in the feature map.
%For effective visualization, we reshaped each selected rows to original spatial size of feature map.
%We observe that it successfully highlights {\color{blue}correlated} regions in the {\color{blue}input} image as well as the {\color{blue}attended pixel,} and generates dense activation maps {\color{blue}based only on} extremely sparse attention. 
We observe that it successfully generates dense activation maps based only on extremely sparse attentions using the correlation of pixels.
It also suggests that using the densified attention as an input to the decoder is more useful to generate accurate and dense segmentation mask than using the original sparse attention.
%which is appropriate to generate accurate and dense segmentation mask in the decoder.
%??? ?? ? ??? ?? ??? ?? ????? ????? highlight ??

\section{Comparisons to Baseline Architectures}
\label{apdx_sec:archs}

\begin{figure*}[!t]
\centering
\includegraphics[width=.97\linewidth] {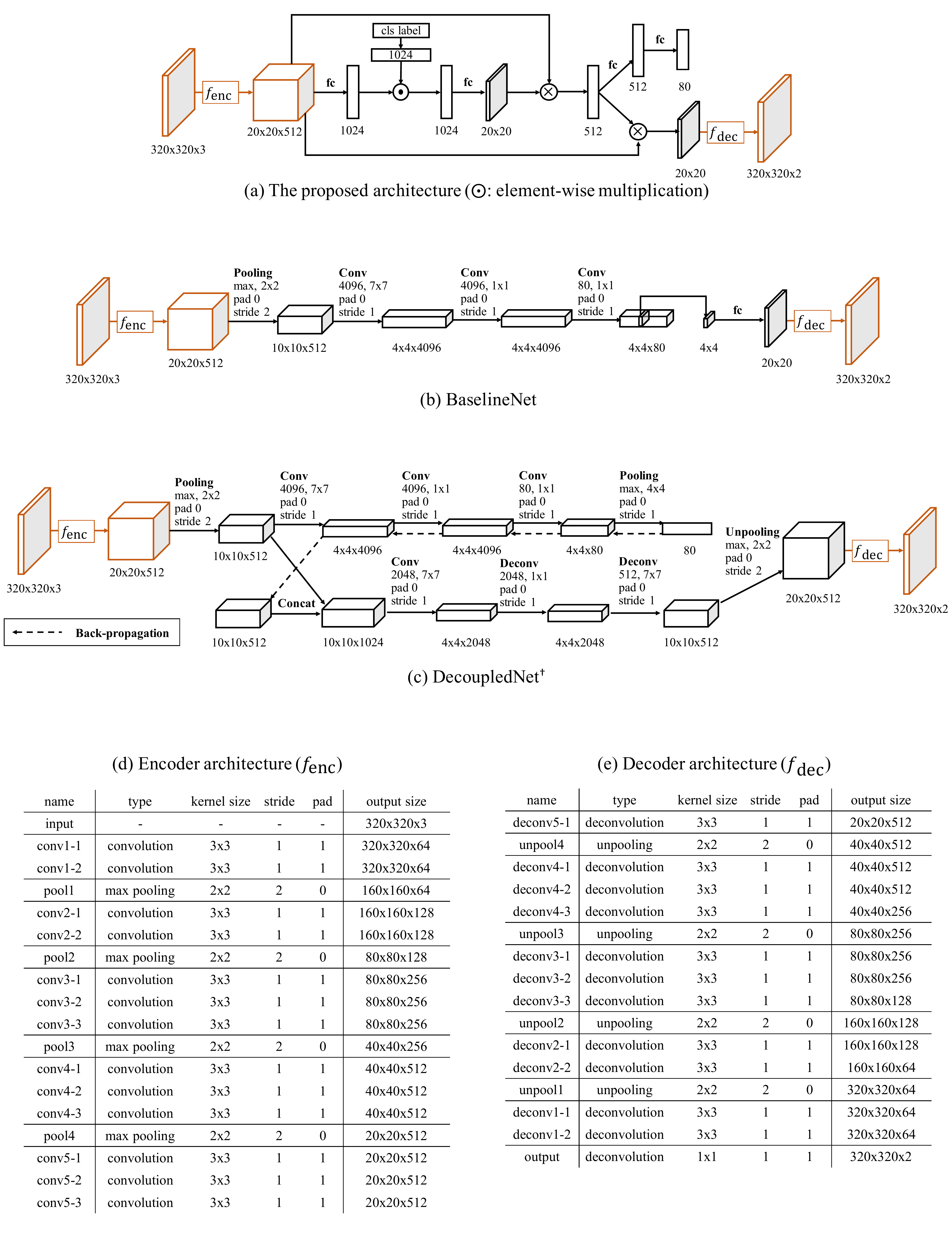}
\caption{
Detailed configurations of the architectures presented in Section~\ref{sec:compare_to_other_baselines}.
}
\label{fig:apdx_archs}
\end{figure*}

Figure~\ref{fig:apdx_archs} illustrates the detailed configurations of the algorithms used in Section~\ref{sec:compare_to_other_baselines}---the proposed algorithm and two baselines denoted by BaselineNet  and DecoupledNet$^\dagger$.
The baseline algorithms are straightforward extensions of existing CNNs for semantic segmentation, which are designed for transfer learning without attention mechanism.
They share the same encoder and decoder architectures with the proposed model but have different approaches to construct the input to the decoder ${\bf s}^l\in\mathrm{R}^M, \forall l \in \mathcal{L}^*$. 
%They share the same encoder (Figure~\ref{fig:archs}.(d)) and decoder  (Figure~\ref{fig:archs}.(e)) architecture with the proposed model but have different mechanisms to convert outputs from the encoder ${\bf A}\in\mathrm{R}^{M\times D}$ to the input to the decoder ${\bf s}^l\in\mathrm{R}^M, \forall l \in \mathcal{L}^*$. 

BaselineNet (Figure~\ref{fig:apdx_archs}{\color{red}(b)}) is an extension of FCN~\cite{Fcn}-style architecture for our transfer learning scenario.
Given an input image, it generates a set of 2D class-specific score maps retaining spatial information based on fully-convolutional network~\cite{Fcn}.
Then for each class presented in the image, it extracts a two-dimensional ($4\times4$) score map of the selected class, and converts it to the input to the decoder through a fully-connected layer. 
Based on this architecture, transfer learning can be achieved by training the decoder in source domain, and fine-tuning the classification network with target domain categories.
DecoupledNet$^\dagger$ (Figure~\ref{fig:apdx_archs}{\color{red}(c)}) is an extension of the decoupled encoder-decoder architecture proposed in \cite{decouplednet} for transfer learning scenario.
%It is composed of separate networks for classification and segmentation.
Given an input image, it first identifies categories presented in the image using the outputs of the classification network, and subsequently generates foreground segmentation of each identified category by the decoder.
To this end, it computes gradient of class score with respect to the feature map by back-propagation, and constructs input to the decoder by combining the feature and gradient maps using a few feed-forward layers.
In this architecture, transfer learning can be achieved by the same way to the BaselineNet; training decoder in source domain, and fine-tuning the classification network with target domain categories.
%Then for each class $l\in\mathcal{L}^*$, it compute gradient of the class score for $l^{th}$ category with respect toand concatenate both the feature and gradients to construct inputs to the decoder, which in turn generates foreground segmentation of $l^{th}$ category.
%Then for each class $l\in\mathcal{L}^*$, it compute gradient of the class score for $l^{th}$ category with respect to the intermediate feature (pool5).
%The gradient map encodes class-specific information, and used to generate foreground segmentation of $l^{th}$ category. 
%To this end, it concatenate both the feature and the gradient, and construct inputs to the decoder by several feedforward layers. 

Table~\ref{tab:voc_result_val} presents comparisons of the proposed algorithm to the baseline architectures.
%Same training data is used to train all algorithms.
DecoupledNet$^\dagger$ directly exploits feature and gradient maps to construct input to the decoder.
Since the both representations are domain-specific and changed by fine-tuning the classification network in different domains, 
the decoder trained on the source domain is difficult to be generalized to unseen categories in the target domain.
Our architecture alleviates this problem using attention, where the attention model provides coherent representation to the decoder across domains.
Compared to BaselineNet, the proposed architecture is more effective to reconstruct crucial information required for segmentation since it employs the attention as well as the original features; it is more appropriate to accomplish more accurate segmentation.

The poor performance of the baseline architectures suggests that transferring segmentation knowledge across domain is a very challenging task, and naive extensions of the existing architectures may not be able to handle this challenge effectively. 
The proposed architecture based on attention mechanism is appropriate to transfer the decoder across domains and obtain accurate segmentation.

%

\iffalse
This section provides detailed descriptions of the architectures for the proposed algorithm and baseline algorithms presented in the main paper.
Two baseline algorithms corresponds to straightforward extension of the existing architectures--BaselineNet from FCN~\cite{Fcn} and DecoupledNet$^\dagger$ from DecoupledNet~\cite{decoupledNet}.
baselineNet and DecoupledNet$^\dagger$.
\fi

\section{Additional Results}
\label{apdx_sec:additional_results}
\begin{figure*}[!t]
\hspace{.8cm}Input Image ~\hspace{.7cm} Ground-truth \hspace{.7cm} EM-Adapt~\cite{Wssl} \hspace{1cm} CCNN~\cite{Ccnn} \hspace{1.1cm}TransferNet \hspace{.6cm} TransferNet+CRF\\\vspace{-0.25cm}\\
%\vspace{-0.2cm}
\centering
\includegraphics[width=0.161\linewidth] {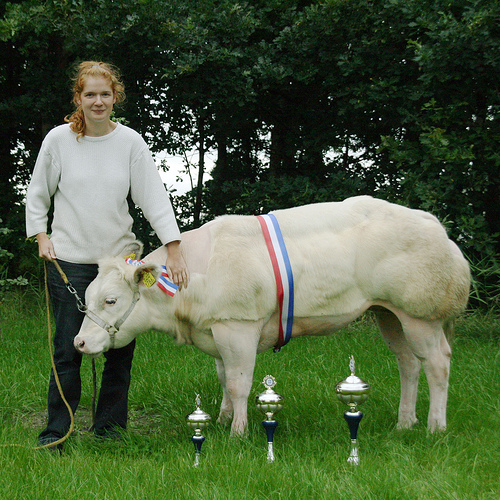}
\includegraphics[width=0.161\linewidth] {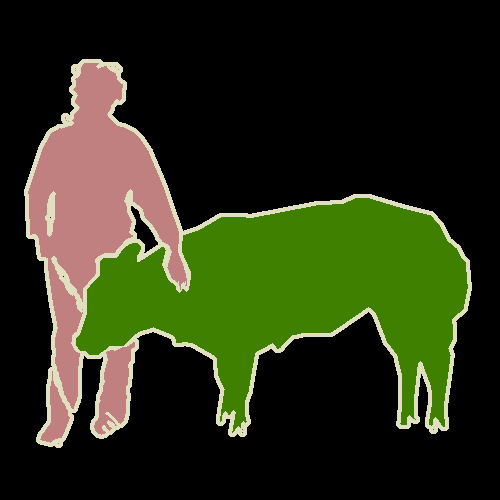}
\includegraphics[width=0.161\linewidth] {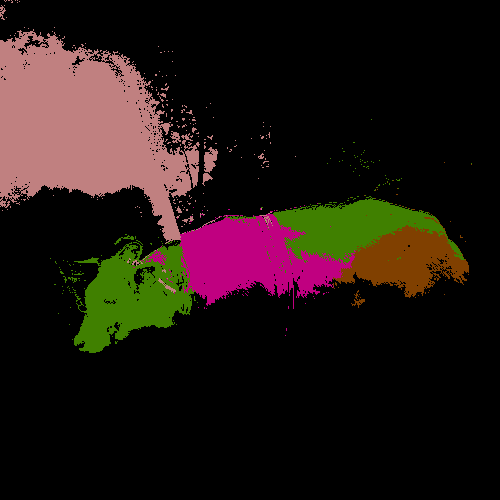}
\includegraphics[width=0.161\linewidth] {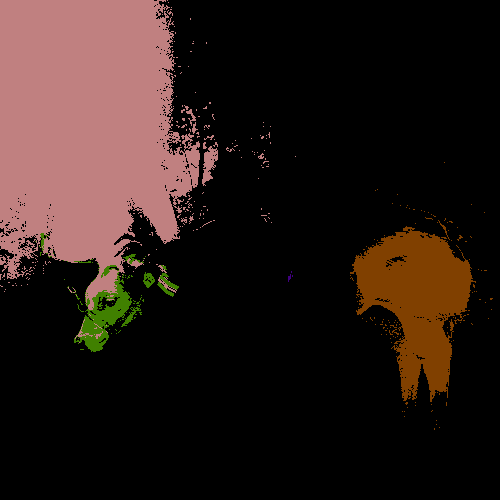}
\includegraphics[width=0.161\linewidth] {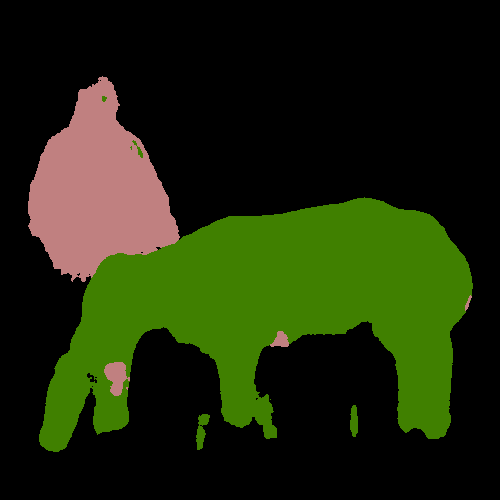}
\includegraphics[width=0.161\linewidth] {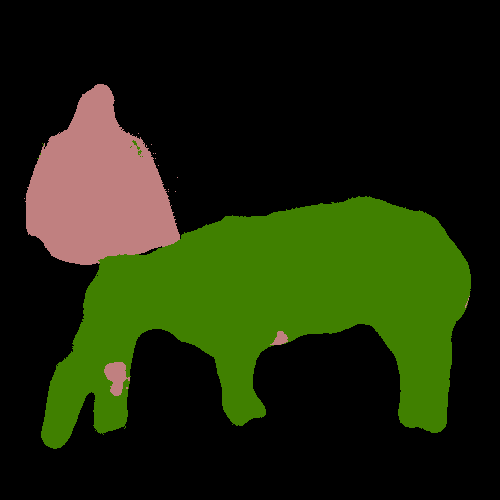}\\
\includegraphics[width=0.161\linewidth] {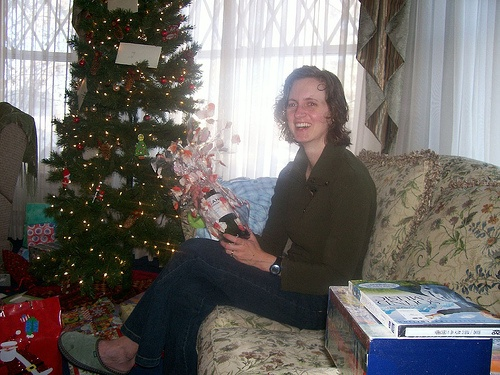}
\includegraphics[width=0.161\linewidth] {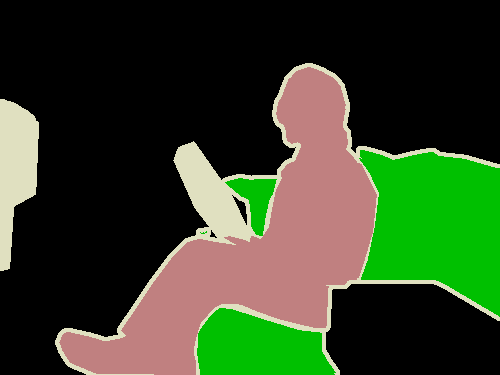}
\includegraphics[width=0.161\linewidth] {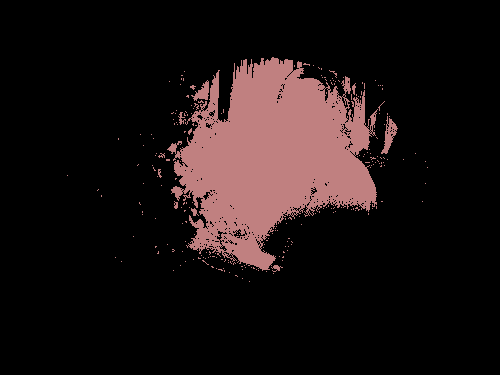}
\includegraphics[width=0.161\linewidth] {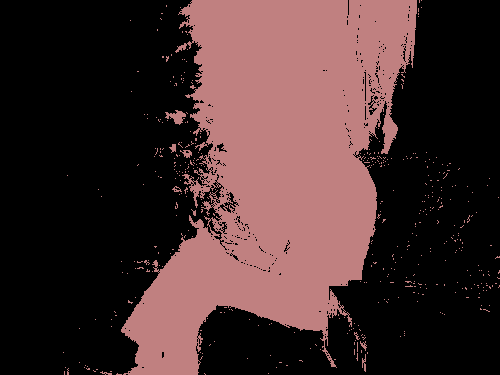}
\includegraphics[width=0.161\linewidth] {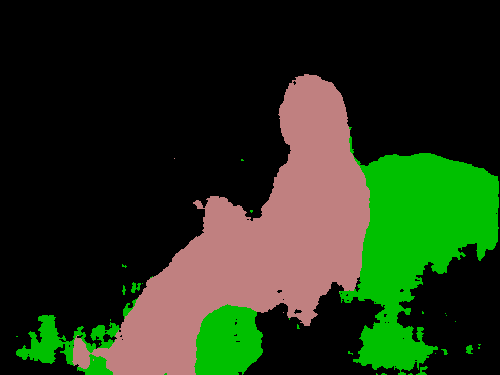}
\includegraphics[width=0.161\linewidth] {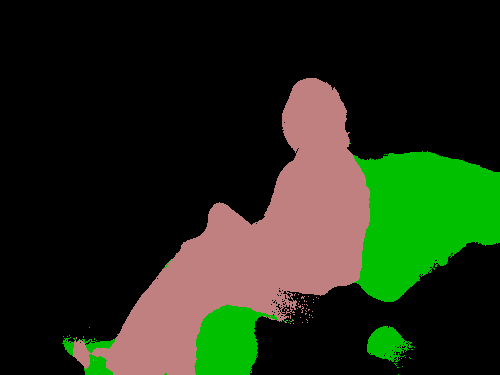}\\
\includegraphics[width=0.161\linewidth] {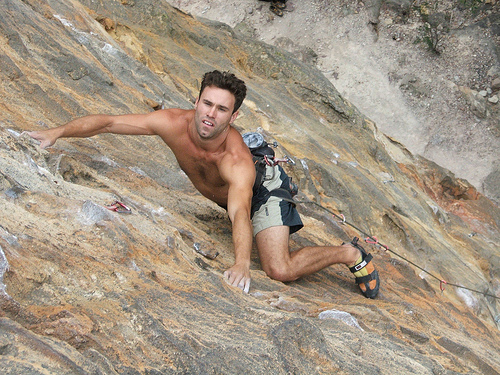}
\includegraphics[width=0.161\linewidth] {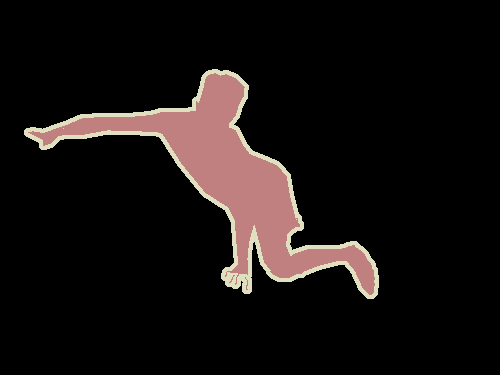}
\includegraphics[width=0.161\linewidth] {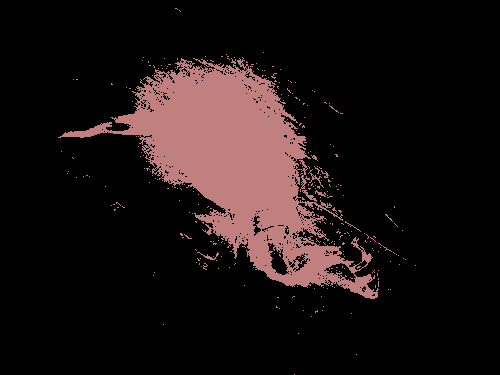}
\includegraphics[width=0.161\linewidth] {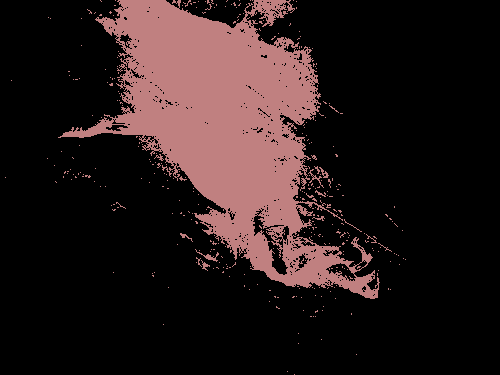}
\includegraphics[width=0.161\linewidth] {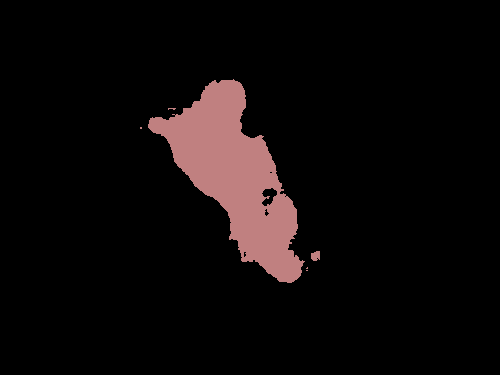}
\includegraphics[width=0.161\linewidth] {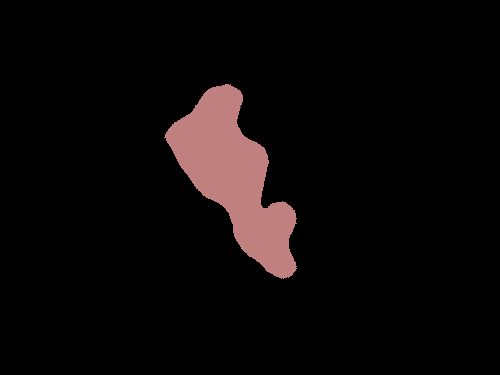}\\
\includegraphics[width=0.161\linewidth] {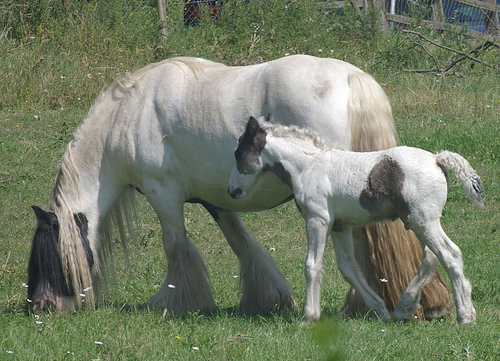}
\includegraphics[width=0.161\linewidth] {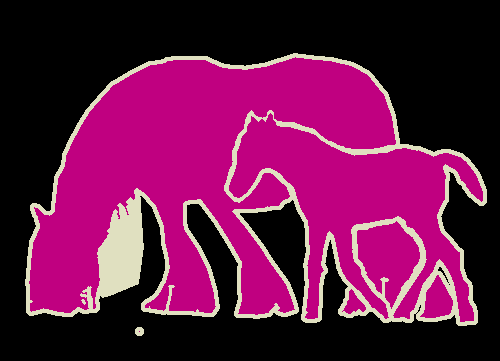}
\includegraphics[width=0.161\linewidth] {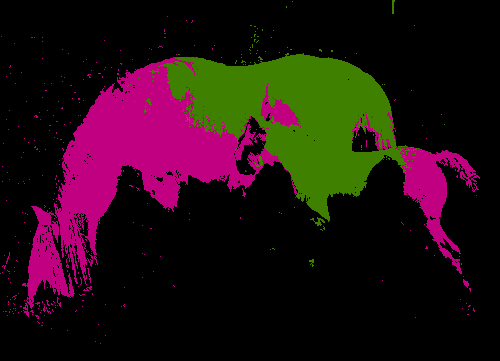}
\includegraphics[width=0.161\linewidth] {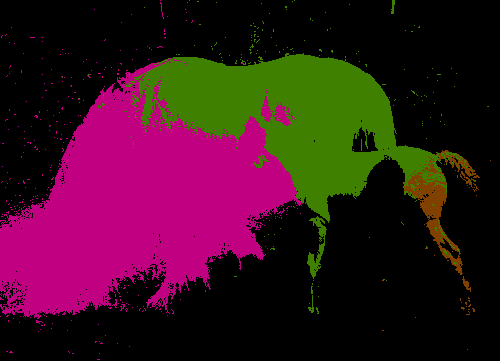}
\includegraphics[width=0.161\linewidth] {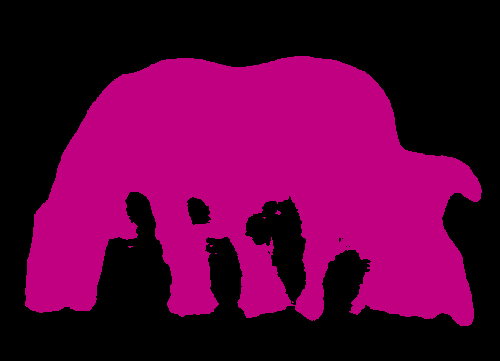}
\includegraphics[width=0.161\linewidth] {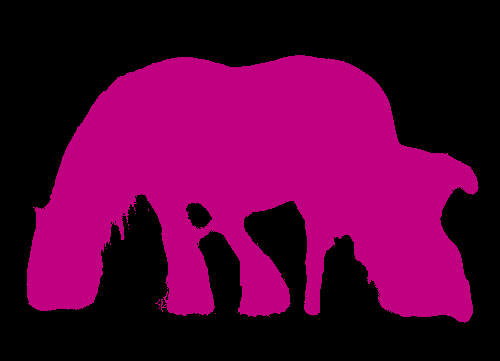}\\
\includegraphics[width=0.161\linewidth] {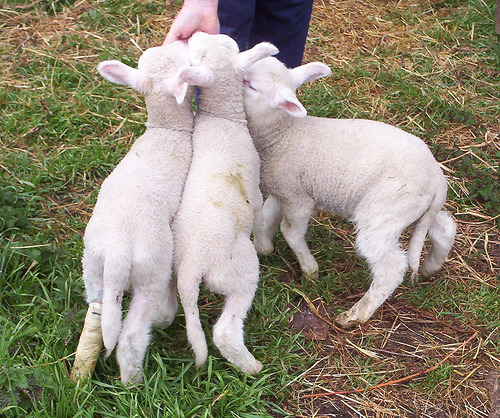}
\includegraphics[width=0.161\linewidth] {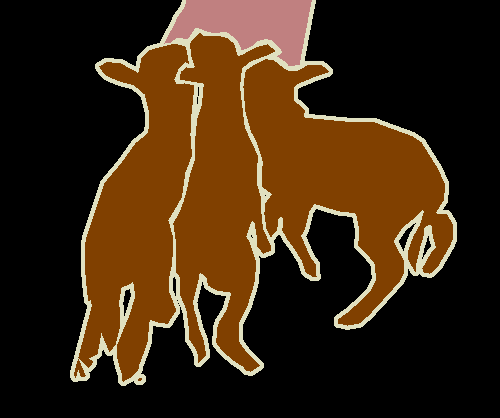}
\includegraphics[width=0.161\linewidth] {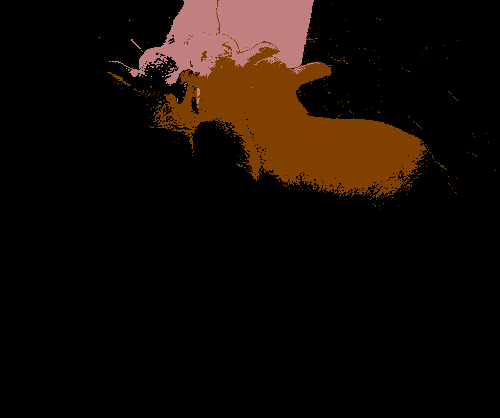}
\includegraphics[width=0.161\linewidth] {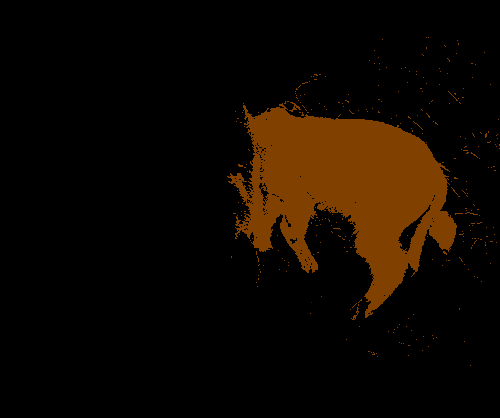}
\includegraphics[width=0.161\linewidth] {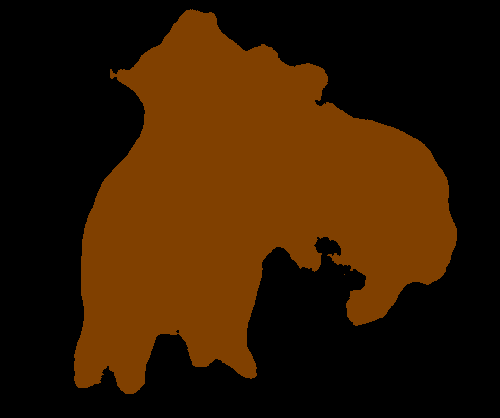}
\includegraphics[width=0.161\linewidth] {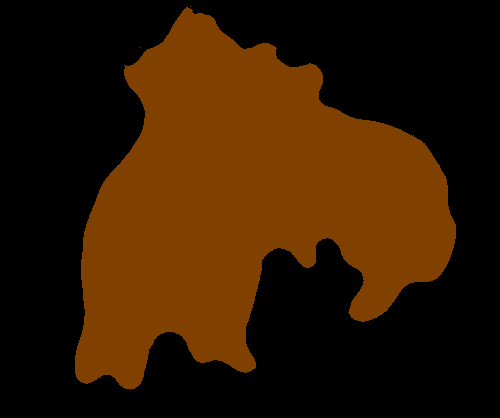}\\
\includegraphics[width=0.161\linewidth] {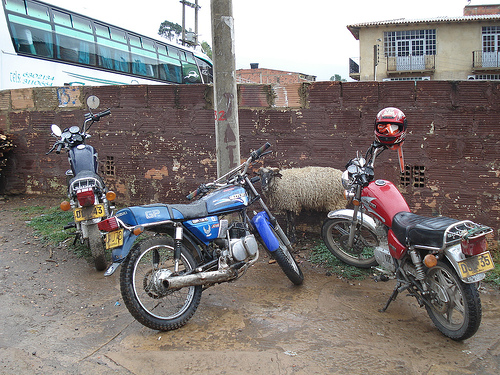}
\includegraphics[width=0.161\linewidth] {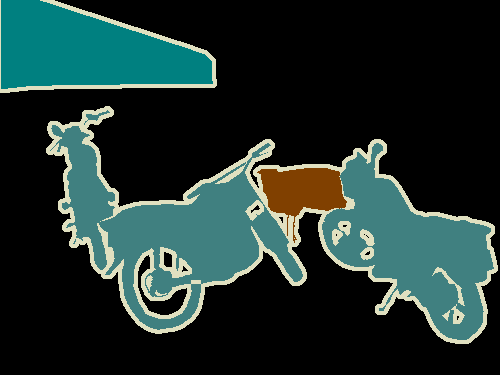}
\includegraphics[width=0.161\linewidth] {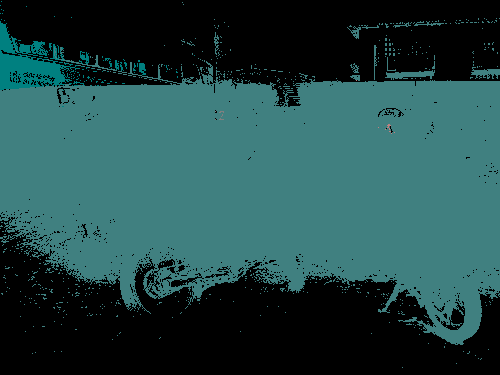}
\includegraphics[width=0.161\linewidth] {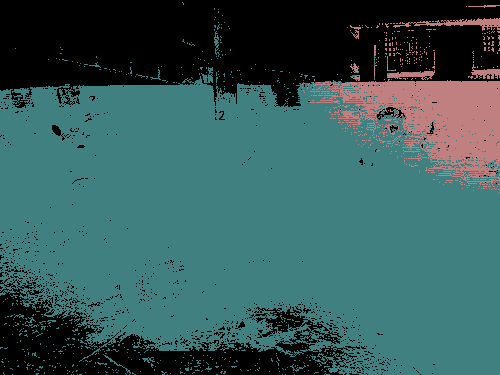}
\includegraphics[width=0.161\linewidth] {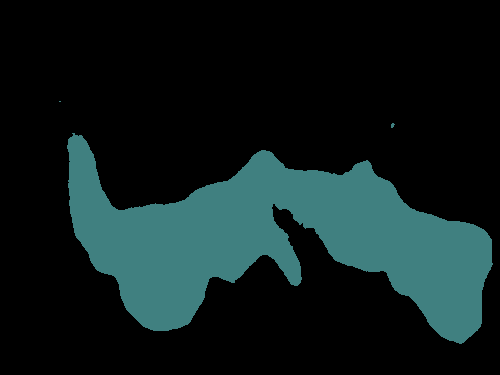}
\includegraphics[width=0.161\linewidth] {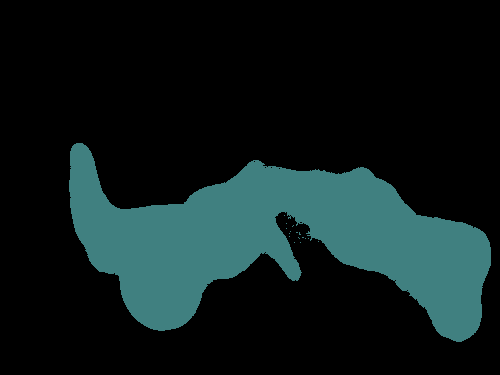}\\
\includegraphics[width=0.161\linewidth] {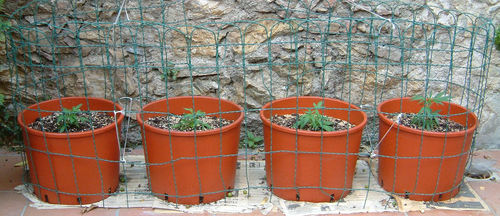}
\includegraphics[width=0.161\linewidth] {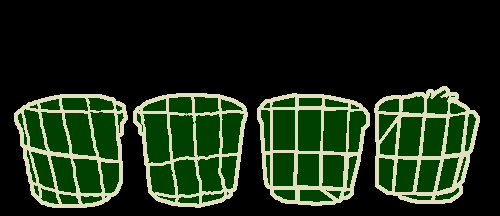}
\includegraphics[width=0.161\linewidth] {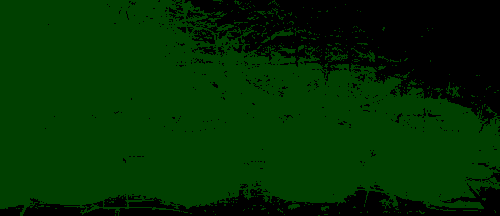}
\includegraphics[width=0.161\linewidth] {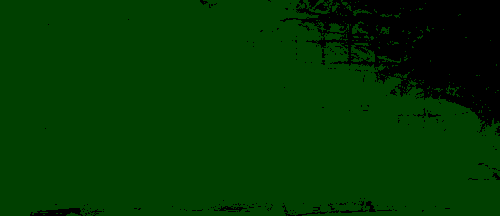}
\includegraphics[width=0.161\linewidth] {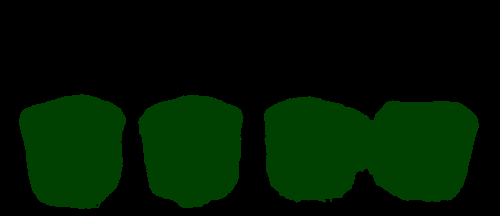}
\includegraphics[width=0.161\linewidth] {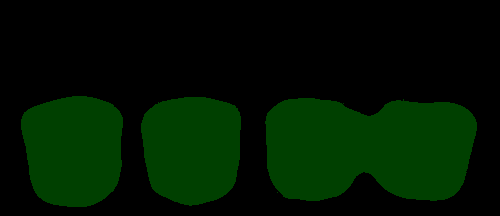}\\
\includegraphics[width=0.161\linewidth] {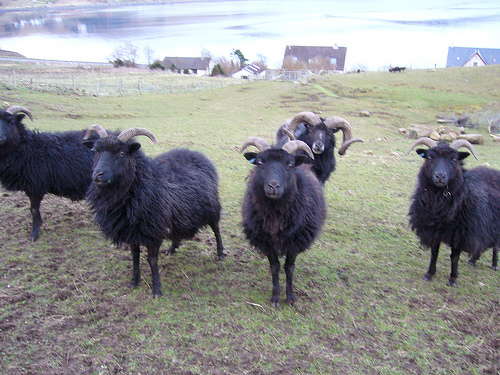}
\includegraphics[width=0.161\linewidth] {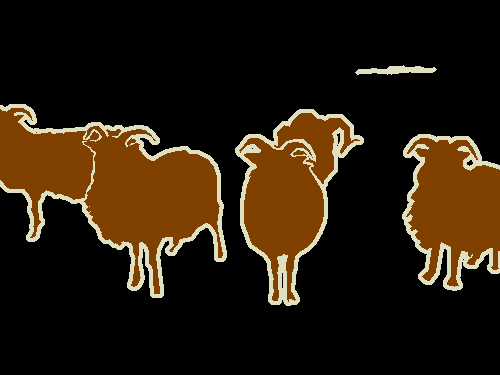}
\includegraphics[width=0.161\linewidth] {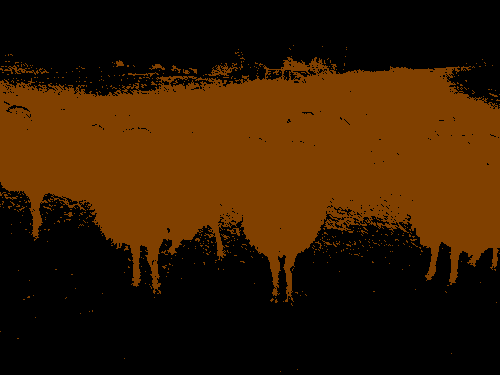}
\includegraphics[width=0.161\linewidth] {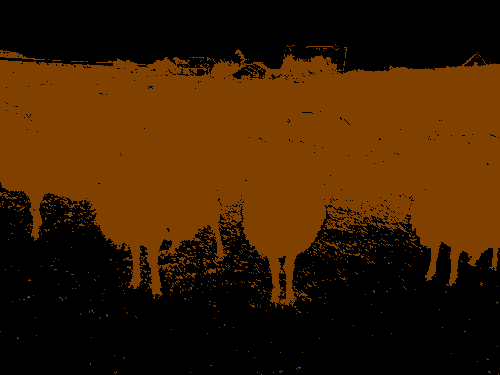}
\includegraphics[width=0.161\linewidth] {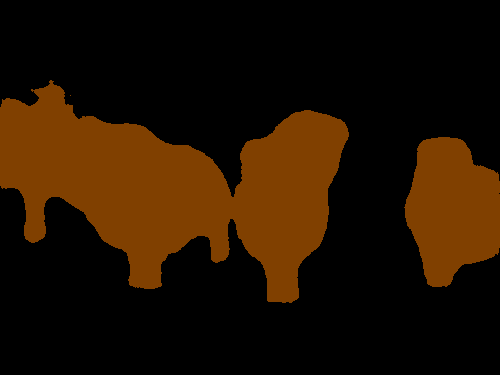}
\includegraphics[width=0.161\linewidth] {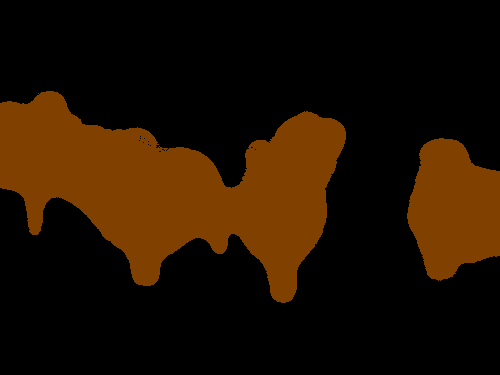}\\
\includegraphics[width=0.161\linewidth] {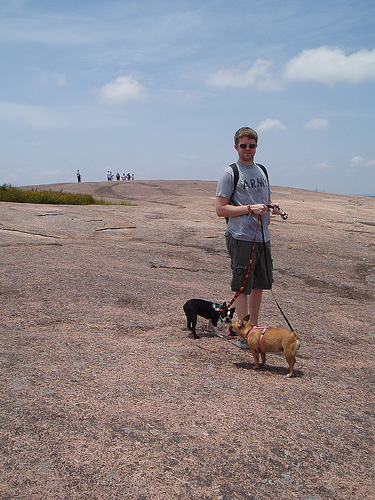}
\includegraphics[width=0.161\linewidth] {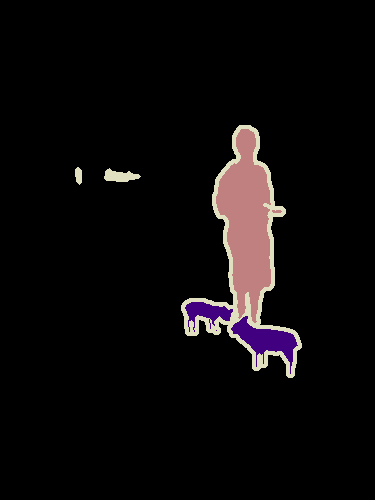}
\includegraphics[width=0.161\linewidth] {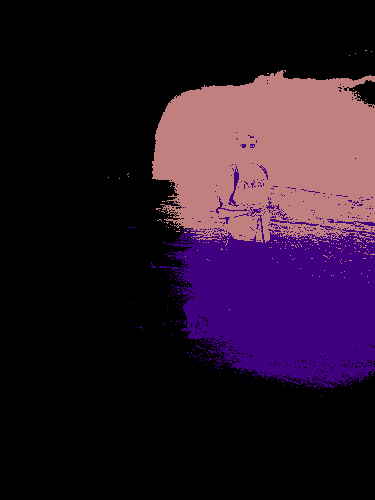}
\includegraphics[width=0.161\linewidth] {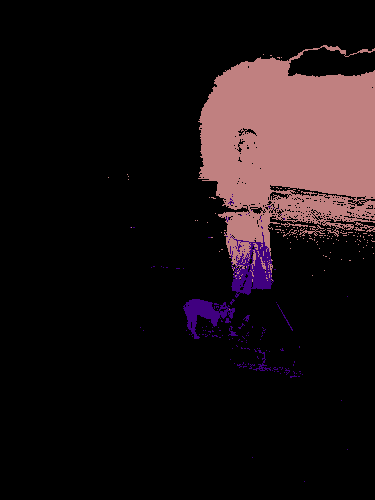}
\includegraphics[width=0.161\linewidth] {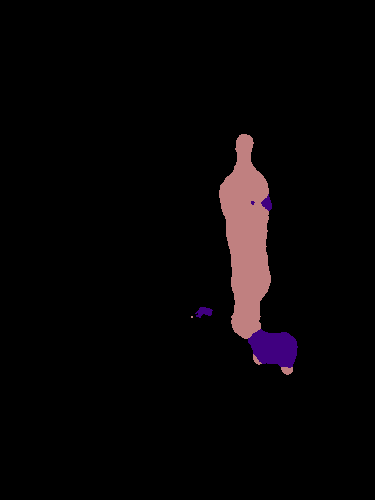}
\includegraphics[width=0.161\linewidth] {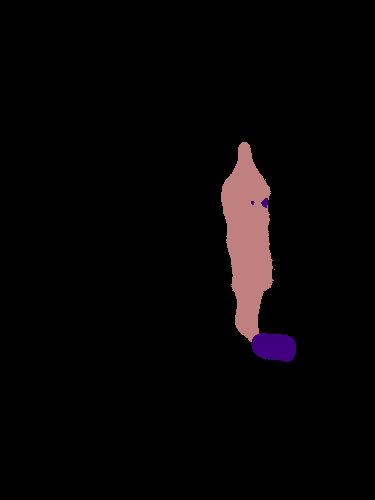}\\
%\end{center}
\caption{
Comparisons of semantic segmentation results on PASCAL VOC 2012 validation images. 
The proposed algorithm tends to find more accurate object boundary compared to other weakly-supervised approaches even without CRF.
}
\label{fig:apdx_qualitative_result}
\end{figure*}

Figure~\ref{fig:apdx_qualitative_result} presents qualitative results of the state-of-the-art weakly-supervised semantic segmentation techniques~\cite{Wssl,Ccnn} including our algorithm on PASCAL VOC 2012 validation images.
%The results show that the compared algorithms heavily rely on post-processing based on CRF to capture object shapes. {\color{red} [This fact is not shown.]}
The compared algorithms adopt post-processing based on fully-connected CRF~\cite{Fullycrf} to refine segmentation results.
Nevertheless, the segmentation of other methods is not successful frequently since the output predictions of the CNN are often too noisy and inaccurate to capture precise object shapes.
Our approach tends to find more accurate object boundaries even without CRF by exploiting the decoder trained with segmentation annotations in different categories, and the results exhibit distinguishing performance compared to existing weakly-supervised approaches.
%The compared algorithms adopt post-processing based on fully-connected CRF, and it is clear that the shape generation heavily rely on the post-processing. 
%Generally, the segmentation quality of the proposed method is substantially better than the compared algorithms.
%Our method tends to find accurate object boundary even without CRF by transferring the decoder trained in the source domain.

%

\end{document}